%% file: main.tex
\documentclass[10pt,twocolumn,letterpaper]{article}

\usepackage[pagenumbers]{iccv} 
\usepackage[accsupp]{axessibility} 
\usepackage{multirow} 
\input{preamble}

\definecolor{iccvblue}{rgb}{0.21,0.49,0.74}
\usepackage[pagebackref,breaklinks,colorlinks,allcolors=iccvblue]{hyperref}

\def\paperID{11138} 
\def\confName{ICCV}
\def\confYear{2025}
\newcommand{\gray}[1]{\textcolor{gray}{#1}}
\newcommand{\green}[1]{\textcolor[RGB]{96,177,87}{#1}}
\newcommand{\fn}[1]{\footnotesize{#1}}
\newcommand{\gbf}[1]{\green{\bf{\fn{(#1)}}}}
\newcommand{\rbf}[1]{\gray{\bf{\fn{(#1)}}}}

\title{Diffusion-Based Imaginative Coordination for Bimanual Manipulation}

\author{
Huilin Xu$^{1}\thanks{Work was done during internship in KAUST.}$  \quad
Jian Ding$^2$ \quad
Jiakun Xu$^{3}$\footnotemark[1] \quad
Ruixiang Wang$^{4}$\footnotemark[1] \quad
Jun Chen$^2$ \quad
Jinjie Mai$^2$ \quad \\
Yanwei Fu$^1$ \quad
Bernard Ghanem$^2$ \quad
Feng Xu$^1$ \quad
Mohamed Elhoseiny$^2$\thanks{Corresponding author: \texttt{mohamed.elhoseiny@kaust.edu.sa}} \\
\\
$^1$Fudan University \quad $^2$King Abdullah University of Science and Technology \\
$^3$ETH Zurich \quad
$^4$The Chinese University of Hong Kong, Shenzhen  
}

\begin{document}
\maketitle
\input{sec/abstract} 
\input{sec/introduction}
\input{sec/related_work}
\input{sec/method}

\input{sec/experiments}


{
    \small
    \bibliographystyle{ieeenat_fullname}
    \bibliography{main}
}

\clearpage

\appendix

\section{Differences between Our Model and GR-1}
 Our model differs fundamentally from GR-1 in both network architecture and training paradigm. Particularly, GR-1 focuses on multi-task generalization via large-scale video pretraining with an autoregressive GPT-style model. In contrast, we target spatio-temporal coordination using a diffusion-based model that jointly predicts future latents and actions without a pretraining stage. Our method also constructs the causal relationship between visual outcomes and action by action-conditioned attention mechanism. Besides, we predict future video latents rather than raw pixels used in GR-1 and show the advantage of latent prediction.

\section{Additional Implementation Details}

\subsection{Implementation Details}
We use DDIM~\cite{song2021denoising} as the noise scheduler with a square cosine schedule \cite{nichol2021improved}, employing 100 diffusion steps during training and 10 steps during inference. Our model predicts the clean sample instead of epsilon. We utilize a pretrained Cosmos tokenizer \cite{agarwal2025cosmos} to extract visual tokens for future frames, using DV $4 \times 8 \times 8$ version with a compression ratio of 256. For ALOHA benchmark, we train for 20, 000 steps with a batch size of 32, while for RoboTwin benchmark, we train for 300 epochs with a batch size of 128. For real-world experiments,  we train for 50 epochs with a batch size of 32. Across all experiments, we adopt a cosine learning rate scheduler with 500 linear warm-up steps to stabilize training.

\subsection{Structure Details}
We provide our architecture and hyperparameter setting details in three evaluate environments, as shown in Table \ref{tab:hyper}. For normalization, we independently scale the minimum and maximum values of each action dimension and each video token dimension to the range \([-1, 1]\). Normalizing actions and tokens to \([-1, 1]\) is essential for DDPM and DDIM predictions, as these models clip their outputs to \([-1, 1]\) to ensure stability \cite{ke20243d}.

\begin{table}
\centering
\resizebox{\linewidth}{!}{
\begin{tabular}{lccc}
\toprule
Model & ALOHA & RoboTwin & Real-World \\
\midrule
Image resolution & 480×640 & 240×320 & 480×640 \\
Backbone & Pretrained ResNet18 & Pretrained ResNet18 & Pretrained ResNet18 \\
\# encoder layer & 4 & 4 &4 \\
\# dncoder layer & 7 & 7 & 7 \\
Chunk size & 100 & 20 & 40 \\
\# Predicted frames & 40 & 20 & 40 \\
Scheduler & DDIM & DDIM & DDIM \\
Prediction type & Sample & Sample & Sample \\
Diffusion steps & 100 & 100 & 100 \\
Diffusion steps (eval) & 10 & 10 & 10 \\
Noise scheduler & Squared\_cosine & Squared\_cosine & Squared\_cosine \\

Tokenizer & Cosmos DV 4×8×8 & Cosmos DV 4×8×8 & Cosmos DV 4×8×8 \\
Patch size & 5 & 5 & 5 \\
\midrule
\textbf{Training}  & ALOHA & RoboTwin & Real-world \\
\midrule
Epochs & 50 & 300 & 50 \\
Batch size & 32 & 256 & 32 \\
Train/Validation ratio & 4:1 & 9:1 & 49:1 \\
Learning rate & 5e-4 & 1e-4 & 5e-5 \\
Lr scheduler & Cosine warmup & Cosine warmup & Cosine warmup \\
Optimizer & AdamW & AdamW & AdamW \\
Prediction weight & 0.2 & 0.2 & 0.2 \\
Image augmentation & RandomShift(15,20) & RandomShift(6,8) & RandomShift(15,20) \\
\bottomrule
\end{tabular}}
\caption{\textbf{Hyperparameters of our method.}}
\label{tab:hyper}
\end{table}

\subsection{Baseline Implementations}
To ensure fair comparison, we report baseline results on both the ALOHA and RoboTwin benchmarks based on either official publications or our own reproductions. For the ALOHA benchmark, we report ACT results directly from the original paper \cite{zhao2023aloha}, and reproduce Diffusion Policy \cite{chi2023diffusion} using its publicly available code and default training settings. For the RoboTwin benchmark, we report the results of Diffusion Policy \cite{chi2023diffusion} and 3D Diffusion Policy \cite{ze20243d_dp} from the RoboTwin paper \cite{mu2024robotwin}. For other baselines, including ACT \cite{zhao2023aloha}, GR-MG \cite{li2025gr_mg}, and RDT-1B \cite{liu2024rdt}, we retrain them using their official implementations and configurations, and evaluate them under the same data and testing protocols as our method. The hyperparameters for ACT and Diffusion Policy are summarized in Table \ref{tab:Hyperparameters of ACT.} and Table \ref{tab:Hyperparameters of DP.}, respectively. 

\begin{table}
    \centering
    \resizebox{\linewidth}{!}{
    \begin{tabular}{l|c|c|c}
    \toprule
         & ALOHA &  RoboTwin & Real-World\\
    \midrule
      Chunk size & 100  & 20  & 40 \\
       Batchsize  & 8 & 256 & 16 \\
      Learning rate &1e-5 & 1e-4 &  2e-5\\
      Lr scheduler & Constant &  Cosine warmup &  Constant \\
     
     Optimizer & AdamW &  AdamW  &  AdamW \\
   \bottomrule
    \end{tabular}}
    \caption{\textbf{Hyperparameters of ACT.}}
    \label{tab:Hyperparameters of ACT.}
\end{table}

\begin{table}
    \centering
    \resizebox{\linewidth}{!}{
    \begin{tabular}{l|c|c|c}
    \toprule
         &  ALOHA & RoboTwin & Real-World\\
    \midrule
      Chunk size  & 100  & 8 & 40 \\
      N\_obs\_step & 2 &3 & 2 \\
      Batchsize  & 20 & 128 & 20 \\
      Learning rate & 1e-4 & 1e-4 & 1e-4\\
      Lr scheduler &  Cosine warmup &  Cosine warmup  & Cosine warmup \\
      Optimizer & AdamW & AdamW &  AdamW \\
   \bottomrule
    \end{tabular}}
    \caption{\textbf{Hyperparameters of Diffusion Policy}.}
    \label{tab:Hyperparameters of DP.}
\end{table}

\subsection{Cosmos Tokenizer}

Cosmos Tokenizer is a core component of the Cosmos World Foundation Model Platform \cite{agarwal2025cosmos}, designed to efficiently transform raw visual data (images and videos) into compact token representations. It supports both continuous and discrete tokenization, preserving spatio-temporal information while reducing computational costs. Designed with a causal architecture, it ensures that token computation depends only on past and current frames, making it well-suited for real-time and sequential tasks. 

For an input video of shape $ (1+T, C, H, W) $, Cosmos Tokenizer compresses it based on the spatial compression factor $ s_{HW} $ and the temporal compression factor $ s_T $, producing an output of shape $ (1 + T/s_T, C, H/s_{HW}, W/s_{HW}) $. The first temporal token represents the first input frame, while subsequent tokens capture temporal dependencies. Spatially, the feature map is downsampled by a factor of $s_{HW} $, resulting in a reduced resolution of $ (H/s_{HW}, W/s_{HW})$.

\begin{figure}
    \centering
    \includegraphics[width=1.05 \linewidth]{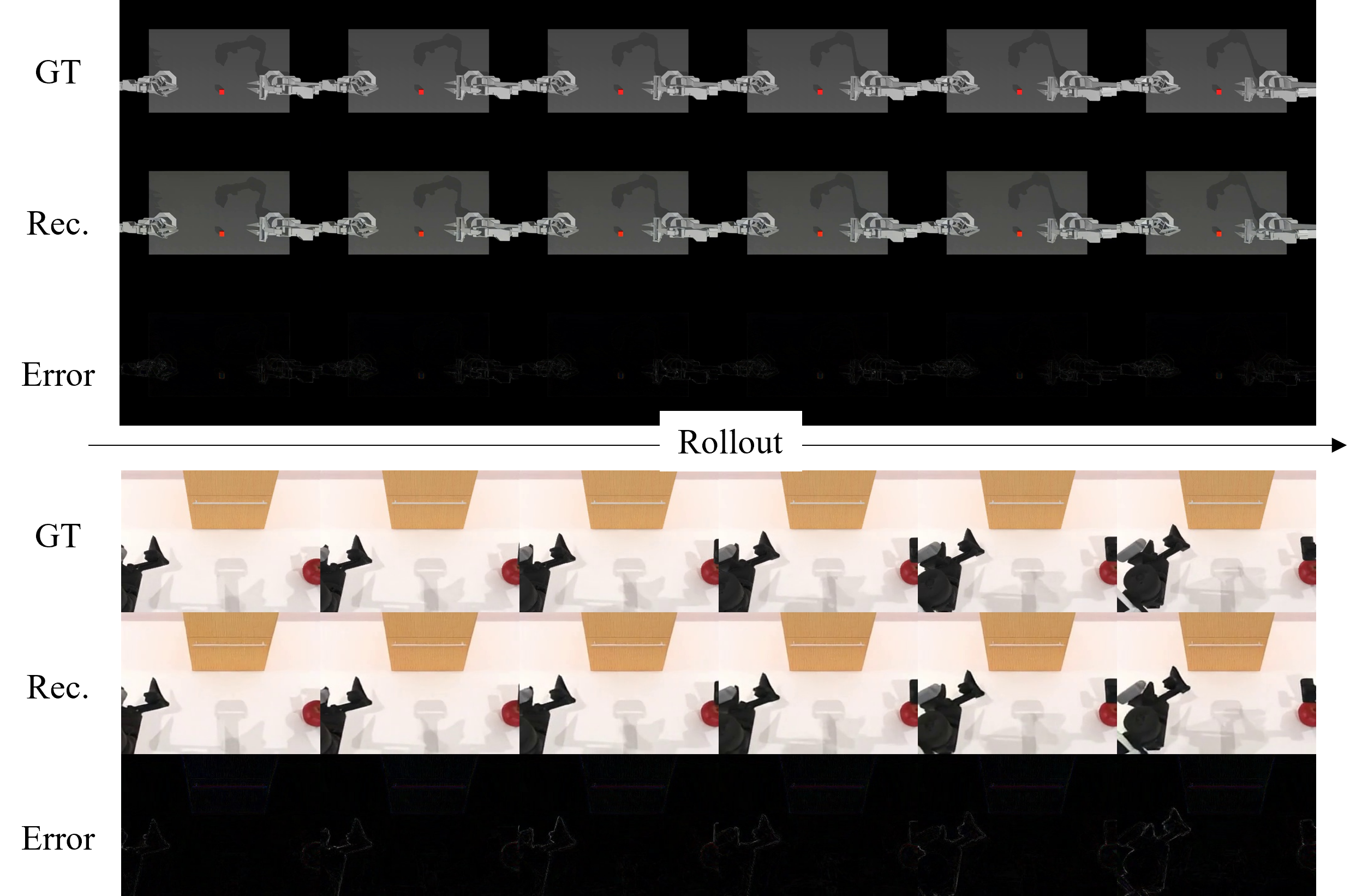}
    \caption{\textbf{Visualization of video reconstruction} using Cosmos Tokenizer on ALOHA \cite{zhao2023aloha} and RoboTwin \cite{mu2024robotwin} benchmark (17 frame sub-clip, DV4x8x8 version).}
    \label{fig:cosmos}
\end{figure}

\cref{fig:cosmos} illustrates video reconstruction of the pretrained Cosmos Tokenizer on two simulated benchmarks. Since the Cosmos tokenizer is trained in various domains such as robotics, driving, egocentric, and web videos, it demonstrates strong generalization capabilities. This makes it a suitable choice as a plug-and-play module for compressing video while retaining information-rich features.

\section{Additional Experimental Results}

\subsection{Comparison with Transformer-Based Baselines}
We compare our method against more transformer-based baselines, including InterACT \cite{leeinteract} and ARP \cite{zhang2024arp}, as shown in Table~\ref{tab:transformer_comparison}. For a fair comparison, we report the original results of InterACT and ARP from their respective papers and retrain our model under the same experimental settings. All results are averaged over 3 random seeds with 50 evaluation episodes each.
Our method achieves the highest average success rate (Avg. SR) of 65.3, consistently outperforming both InterACT and ARP.
\begin{table}
\centering
\renewcommand{\arraystretch}{1.0} 
\resizebox{\linewidth}{!}{
\begin{tabular}{@{}cccc@{}}
\toprule
Method & Avg. SR $\uparrow$ & Transfer Cube (Human) & Insertion (Human) \\
\midrule
InterACT \cite{leeinteract} & \underline{63.0}  & 82 & \underline{44} \\
ARP \cite{zhang2024arp}      & 59.4  & \textbf{94} & 24.8 \\
Ours     & \textbf{65.3} & \underline{84} & \textbf{46.7} \\
\bottomrule
\end{tabular}
}
\caption{\textbf{Comparison with more transformer-based methods.}}
\label{tab:transformer_comparison}
\end{table}

\subsection{Results in Multi-task setting}
To evaluate the effectiveness of our method in more complex multi-task scenarios, we extend our single-task framework to a language-conditioned multi-task policy, inspired by prior work~\cite{roboagent}. Specifically, we incorporate FiLM-based conditioning~\cite{perez2018film} to inject task descriptions into the policy. We construct a multi-task benchmark using three representative bimanual tasks from RoboTwin and compare our method against RDT-1B~\cite{liu2024rdt}. As shown in Table~\cref{tab: performance on multi-task setting}, our approach consistently outperforms RDT-1B, highlighting its strong ability in complex dual-arm manipulation.

\begin{table}[t]
\centering
\resizebox{\linewidth}{!}{
\begin{tabular}{@{}lllll@{}}
\toprule
& Avg. $\uparrow$ & Diverse & Blocks  & Put Apple  \\
Method & Success (\%) $\uparrow$ & Bottle Pick &Stack (Easy) &  Cabinet \\
\midrule
RDT-1B \cite{liu2024rdt} & 54.0 & 14 & \textbf{66} & 82 \\
Ours   & \textbf{62.0 \textcolor[rgb]{0.3, 0.7, 0.3}{(+8.0)} } & \textbf{24} & 62 & \textbf{100} \\
\bottomrule
\end{tabular}}
\caption{\textbf{Performance comparison under multi-task setting.} We extend our methods to a multi-task policy and beats RDT-1B \cite{liu2024rdt}.}
\label{tab: performance on multi-task setting}
\end{table}

\subsection{Visualization of Video Prediction}
\begin{figure}
    \centering
    \includegraphics[width=0.95\linewidth]{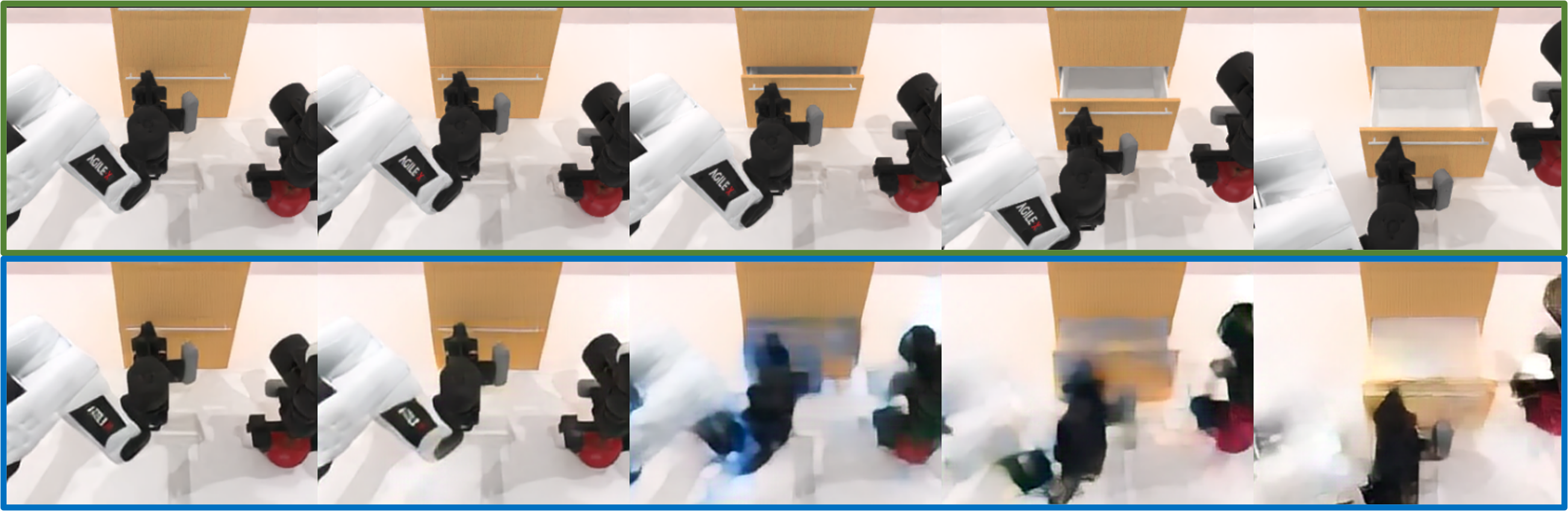}
    \caption{\textbf{Visualization of video prediction results.}}
    \label{fig:pred_len_plot}
\end{figure}

We visualize predicted frames in \cref{fig:pred_len_plot}. While not photorealistic, our method generates semantically meaningful predictions (e.g., drawer opening) that reflect task-relevant dynamics.  We respectfully clarify that our model is not designed to generate photorealistic frames, but rather to capture key task dynamics (e.g., drawer opening) for robotic action prediction. As prior work has shown that accurate modeling of task-relevant latent dynamics is more important than pixel-level reconstruction for effective control, e.g. TD-MPC2 \cite{hansen2023tdMPC2}, MPI \cite{MPI}. We do not explicitly optimize for video quality, as our objective is not visual fidelity but task-relevant dynamics

\subsection{Detailed Result on Data Efficiency Setting}
\begin{table*}
    \centering
    \resizebox{\linewidth}{!}{
    \begin{tabular}{@{}c c c c c c  c c c c @{}}
        \toprule
         & \cellcolor{gray!30} Avg. & Block & Block & Blocks  & Blocks & Bottle & Container & Diverse  & Dual Bottles   \\
        
        Method &\cellcolor{gray!30} Success $\uparrow$  & Hammer  Beat & Handover & Stack (Easy) & Stack (Hard) & Adjust & Place & Bottles Pick & Pick (Easy)  \\
        \midrule
        3D Diffusion Policy \cite{ze20243d_dp} & \cellcolor{gray!30}  \textbf{30.4} & 55.7 & \textbf{89}    & - & - & \textbf{64.7}  & \textbf{52.7} & \textbf{11.3}  & \textbf{40.3 }\\
        Diffusion Policy \cite{chi2023diffusion} & \cellcolor{gray!30}  1.5 & 0.0  & 0.0 & 0.0 & 0.0 & 6.3 & 1.7 & 0.7 & 1.7  \\
        ACT \cite{zhao2023aloha} & \cellcolor{gray!30}  15.4 & 45.3 & 65.67 & 3.67 & 0.0 & \underline{38.33}  & 9.67 & 0.7 & 27.7  \\
        Ours & \cellcolor{gray!30}  \underline{27.3} & \textbf{60.33} & \underline{88.33} & \textbf{4.33} & \textbf{0.33} & \underline{38.33}  & \underline{40.33} & \underline{1.33} & \underline{32.33}  \\
        
        \midrule
          &  Dual Bottles 
         & Dual Shoes  & Empty Cup & Mug  Hanging   & Mug  Hanging & Pick Apple  & Put Apple & Shoe & \cellcolor{gray!30} Coord. \\  
        Method  &   Pick (Hard) 
         & Place & Place &(Easy) & (Hard) &  Messy &  Cabinet & Place  & \cellcolor{gray!30}  Avg. $\uparrow$  \\  %
        \midrule
        3D Diffusion Policy  \cite{ze20243d_dp} &  \underline{31.7} & \textbf{4.0} & \textbf{33.7} & \textbf{7.3} & \textbf{4} & \underline{4}  & \underline{50.0} & \textbf{38} &  \cellcolor{gray!30} \underline{25.1} \\
        Diffusion Policy  \cite{chi2023diffusion}& 8.0 & 0.0 & 0.0 & 0.0 & 12.0 & 5.3  & 0.0 & 0.0 &  \cellcolor{gray!30} 0.0 \\
        ACT \cite{zhao2023aloha} & 17.0 & 2.7 & 3.0 & 0.0 & 0.0 & 7.0  & 14.33 & 12 &  \cellcolor{gray!30} 13.9 \\
        Ours & \textbf{34.33} & \underline{3.67} & \underline{16.67} & \underline{0.67} & 1.0 & \textbf{8.67}  & \textbf{76.0} & \underline{29.33} &  \cellcolor{gray!30} \textbf{28.4} \\
        \bottomrule
    \end{tabular}}
    \caption{\textbf{Evaluation on data efficiency setting.} We report the mean of success rates averaged over 3 random seeds.  Best score in \textbf{bold}, second-best \underline{underlined}. Coord. Avg. denotes the averaged success rate of tasks in the coordinated subset.} %
    \label{tab: date_effiency}
\end{table*}

We evaluate methods under the data efficiency setting, shown in Table \ref{tab: date_effiency}. The performance of all methods degrades compared to the default setting due to the limited training data. Among the baselines, 3D Diffusion Policy achieves the highest overall success rate, benefiting from its 3D point cloud representation, which has been shown to exhibit strong sample efficiency. Our method achieves the best result among all 2D policy models. In \textit{Seq-coordinate} tasks, our method even outperforms 3D DP, demonstrating its effectiveness in capturing sequential dependencies on the data efficiency setting.

\subsection{Impact of Video Token Type}
We also study the impact of various video tokens from Cosmos-Tokenizer\cite{agarwal2025cosmos}. As shown in~\cref{tab:albation_results_video_token}, two different video tokens, discrete tokens and continuous tokens are implemented based on our model and the best results are reported for both tokens with same training seed. Our empirical analysis demonstrates that discrete tokens outperform continuous tokens in action prediction. Notably, this finding contrasts with results from video prediction studies, indicating that the underlying factors merit further investigation in future research.

\begin{table}[h]
    \centering
    \resizebox{\linewidth}{!}{
    \begin{tabular}{@{}c c c c c c@{}}
        \toprule
         Video & \cellcolor{gray!30} Avg. &  \multicolumn{2}{c}{Transfer Cube} & \multicolumn{2}{c}{Insertion}  \\
         Token  & \cellcolor{gray!30} Success $\uparrow$ & Scripted & Human & Scripted & Human \\
        \midrule
         DV & \cellcolor{gray!30} 73.7 & 98 & 78 & 79 & 40 \\
         CV & \cellcolor{gray!30} 70.7 & 95 & 77 & 86 & 25  \\
        \bottomrule
    \end{tabular}}
    \caption{\textbf{Ablation of video token.} DV denotes discrete video token and CV denotes continuous video token.}
    \label{tab:albation_results_video_token}
\end{table}


\section{Additional Task Details}

\subsection{Task Description}

\begin{figure*}
    \centering
    \includegraphics[width= 0.9 \linewidth]{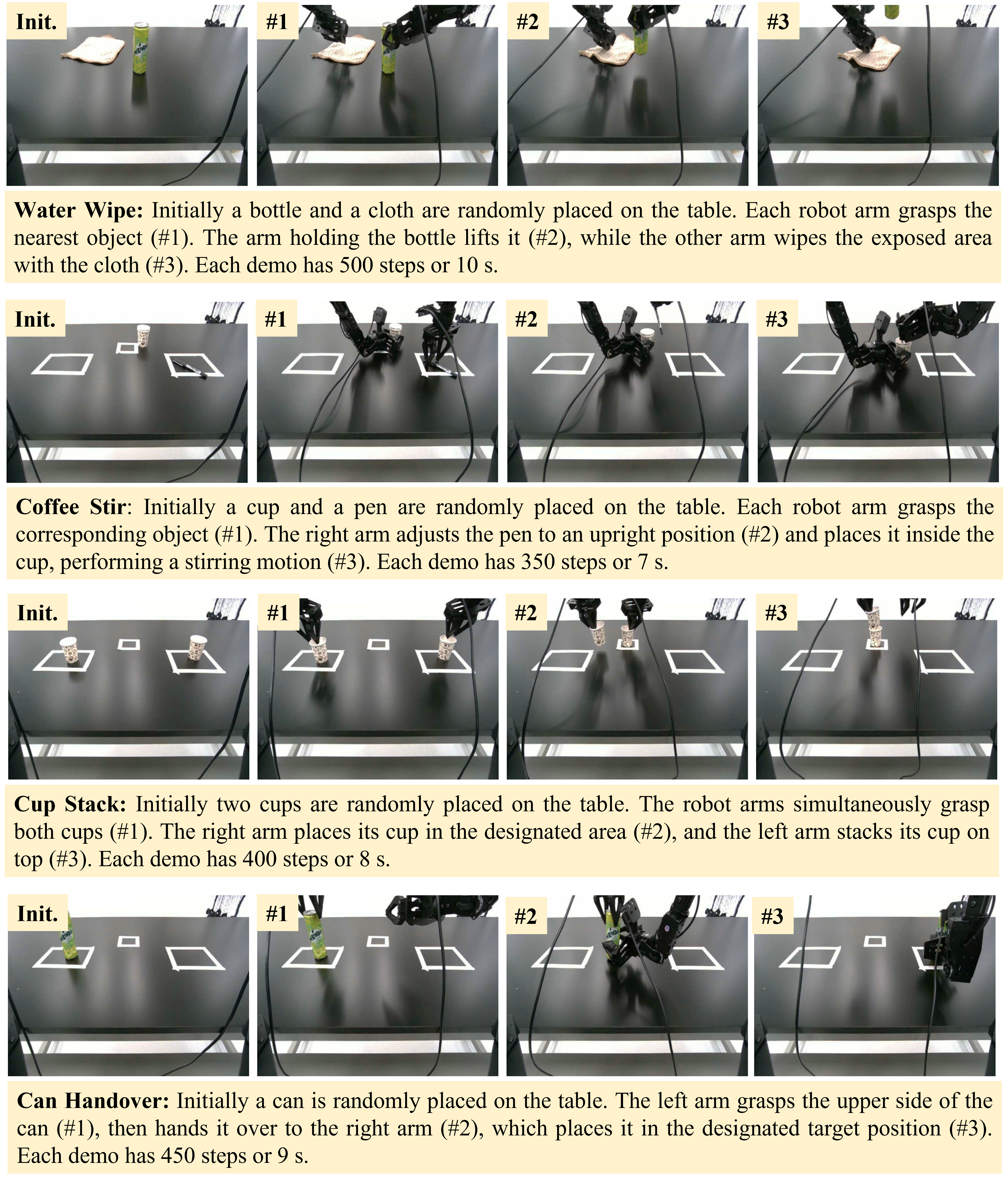}
    \caption{\textbf{Task definition of real-world experiments.}}
    \label{fig:vis_real_world}
\end{figure*}

 For details of tasks in the ALOHA \cite{zhao2023aloha} and RoboTwin \cite{mu2024robotwin} simulation benchmarks, please refer to their original papers. Table \ref{tab:tasks_dis} provides a comprehensive overview of the real-world benchmark robot tasks, illustrated in Fig. \ref{fig:vis_real_world}.

\begin{table}[h]
\centering
\resizebox{\linewidth}{!}{
\begin{tabular}{ccp{4.4cm}}
\toprule
         Task & \#Steps & Task Description \\
\midrule
Water Wipe & 500 & Lift a bottle and wipe the exposed area with a cloth. \\
Coffee Stir & 350 &Pick up a cup and a pen, then stir inside the cup. \\
Cup Stack & 400 & Grasp both cups, place the right one first, then stack the left. \\
Can Handover & 450 &The left arm hands a can to the right arm, which places it. \\
\bottomrule
\end{tabular}}
\caption{\textbf{Real-World task descriptions}}
\label{tab:tasks_dis}
\end{table}
 

\subsection{Task Category}

We categorize the bimanual tasks in RoboTiwn benchmark into three types:

\begin{itemize}
    \item \textbf{Dominant-select} – The executing arm is chosen based on the object’s position. Tasks include:
    \begin{itemize}
        \item \textit{Block Hammer Beat, Empty Cup Place, Pick Apple Messy, Shoe Place, Bottles Ajust, Container Place}

    \end{itemize}
    
    \item \textbf{Sync-bimanual} – Both arms operate independently but simultaneously. Tasks include:
    \begin{itemize}
 
        \item \textit{Diverse Bottle Pick, Dual Bottles Pick (Easy), Dual Bottles Pick (Hard), Dual Shoes Place }

    \end{itemize}

    \item \textbf{Seq-coordinate} – Tasks require sequential coordination with temporal dependencies. Tasks include:
    \begin{itemize}
        \item \textit{Block Handover, Blocks Stack (Easy), Blocks Stack (Hard), Mug Hanging (Easy), Mug Hanging (Hard), Put Apple Cabinet}
    \end{itemize}
\end{itemize}

\end{document}

%% file: preamble.tex
\usepackage{colortbl}
\usepackage{bbding} 

\usepackage{tikz}
\usepackage{pgfmath}
\usepackage{ifthen}
\usetikzlibrary{matrix,
                positioning,
                calc,
                arrows.meta,
                shapes,
                patterns,
                fit,
                decorations.pathmorphing,
                ext.patterns.images}
\usepackage{pgfplots}
\pgfplotsset{compat=1.18}
\usepgfplotslibrary{groupplots}
\pgfplotsset{every tick label/.append style={font=\small}}
\pgfsetupimageaspattern[width=4cm]{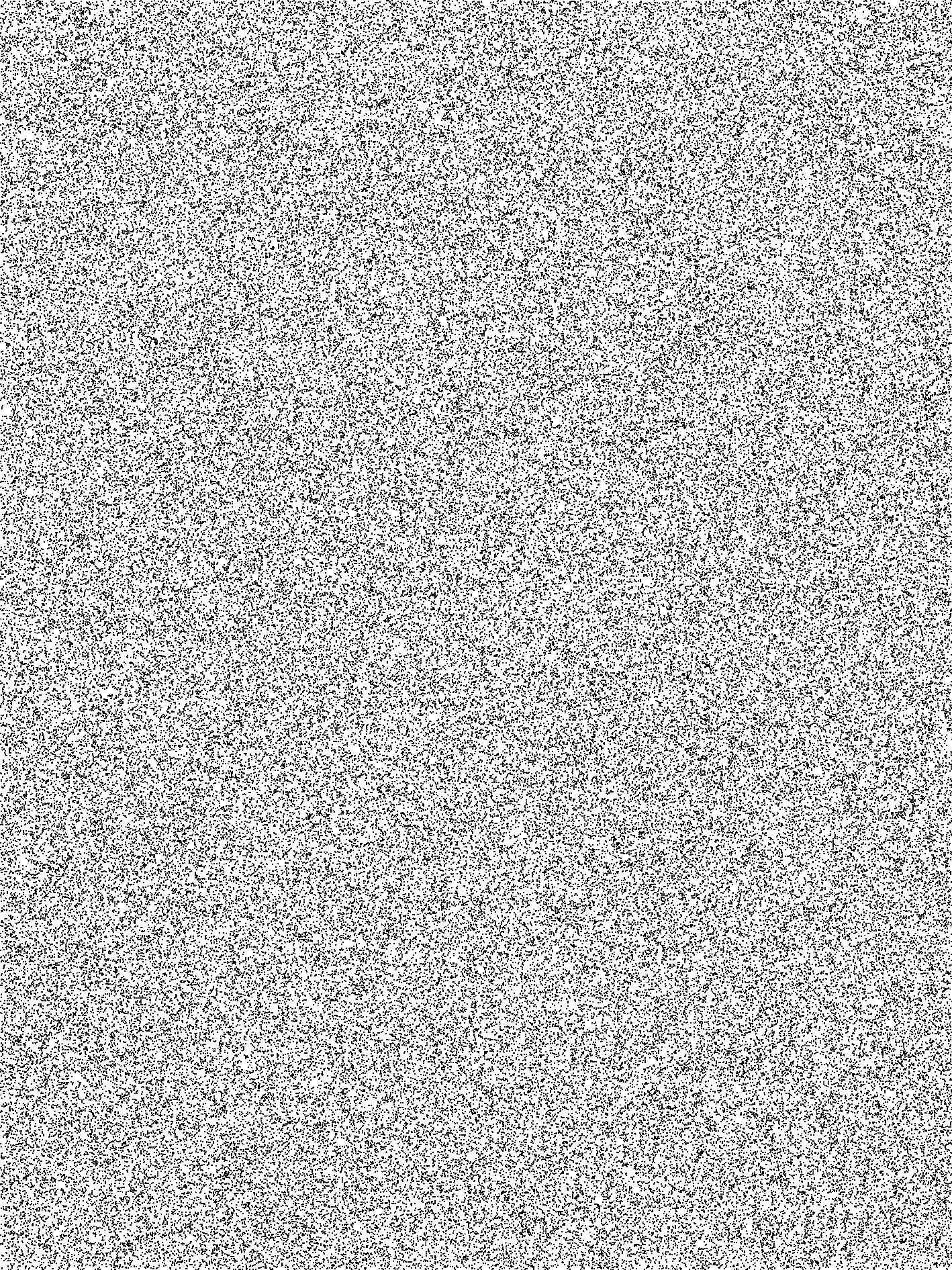}{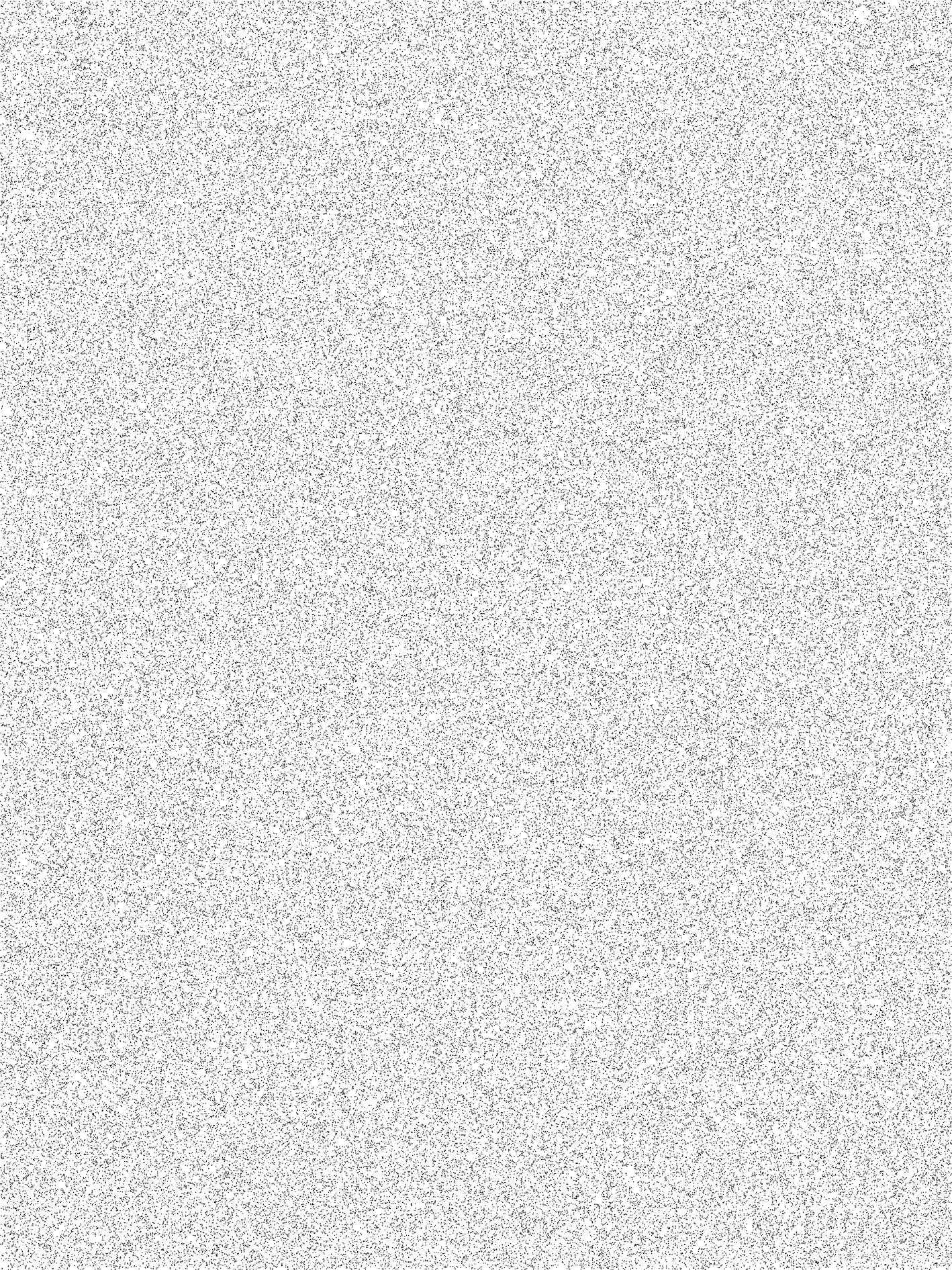}
\pgfsetupimageaspattern[width=4cm]{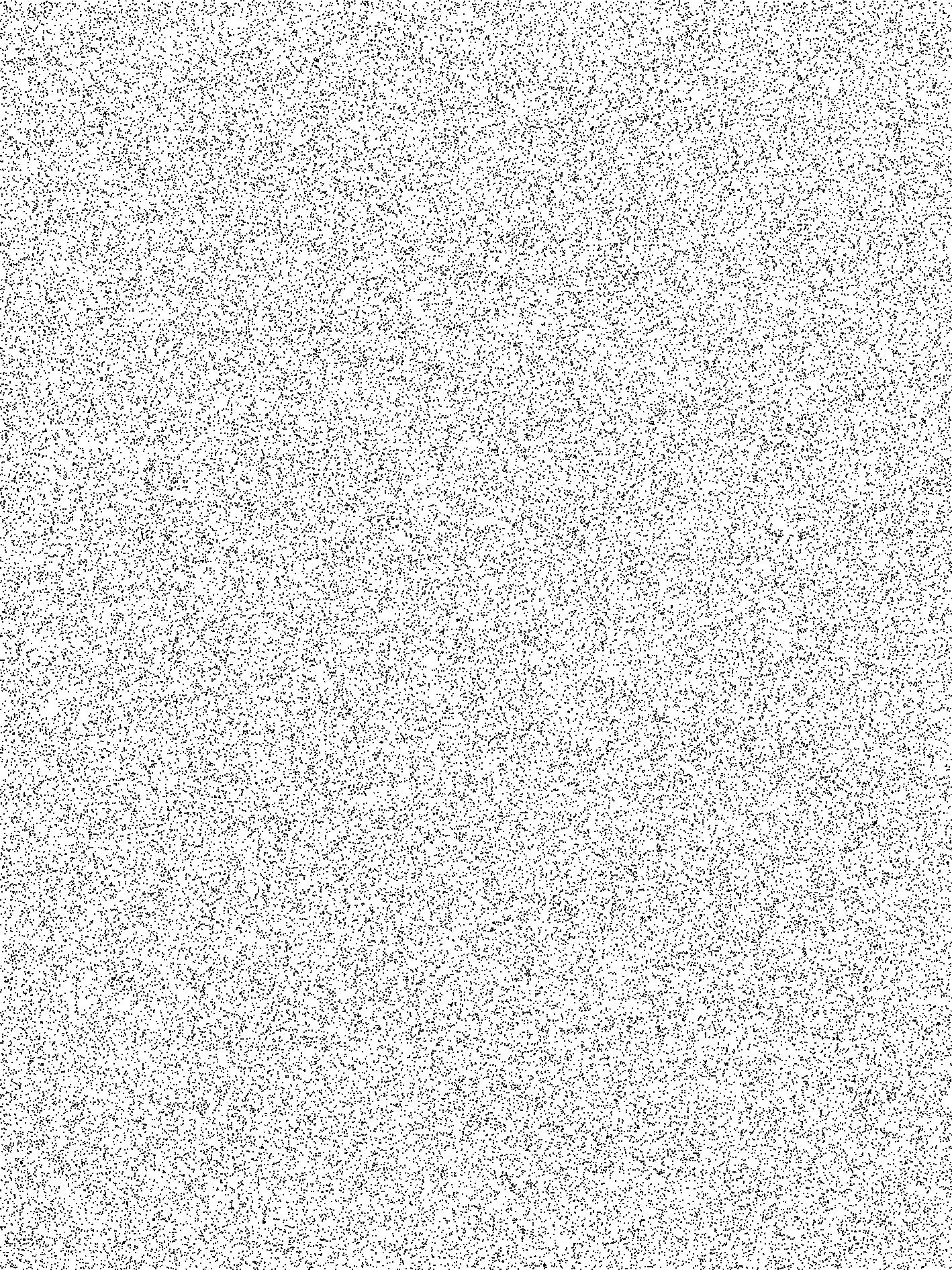}{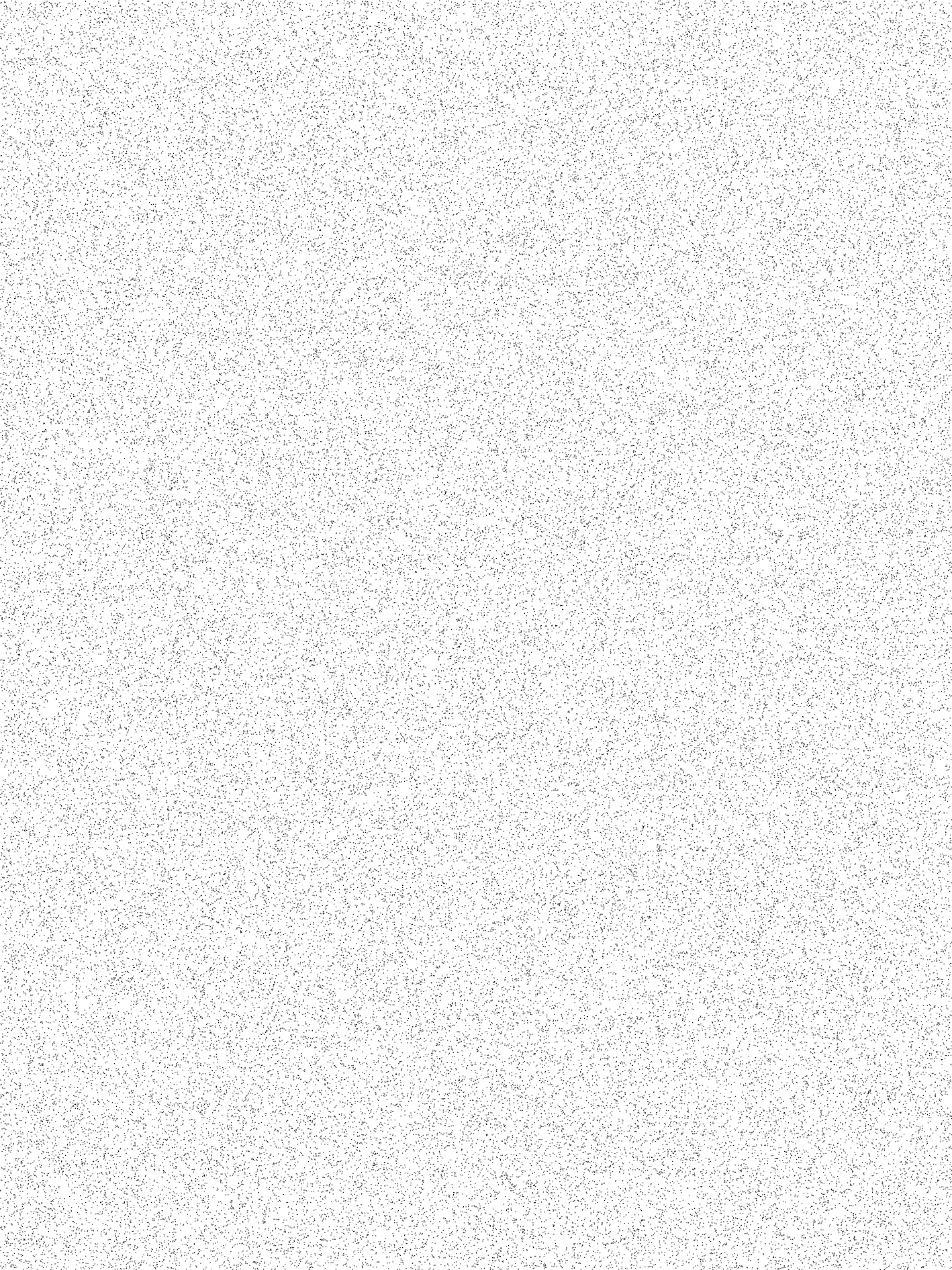}
\pgfsetupimageaspattern[width=4cm]{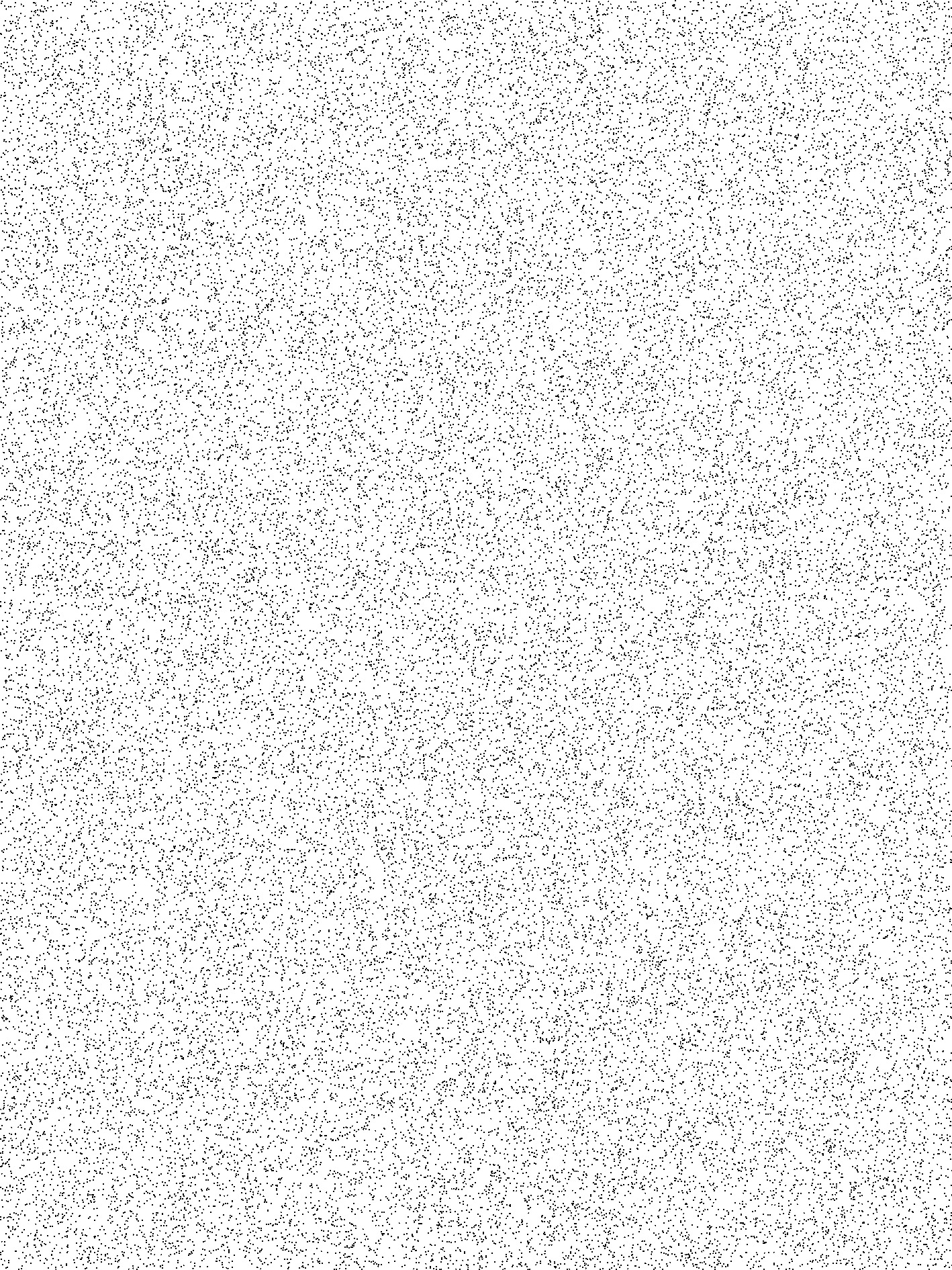}{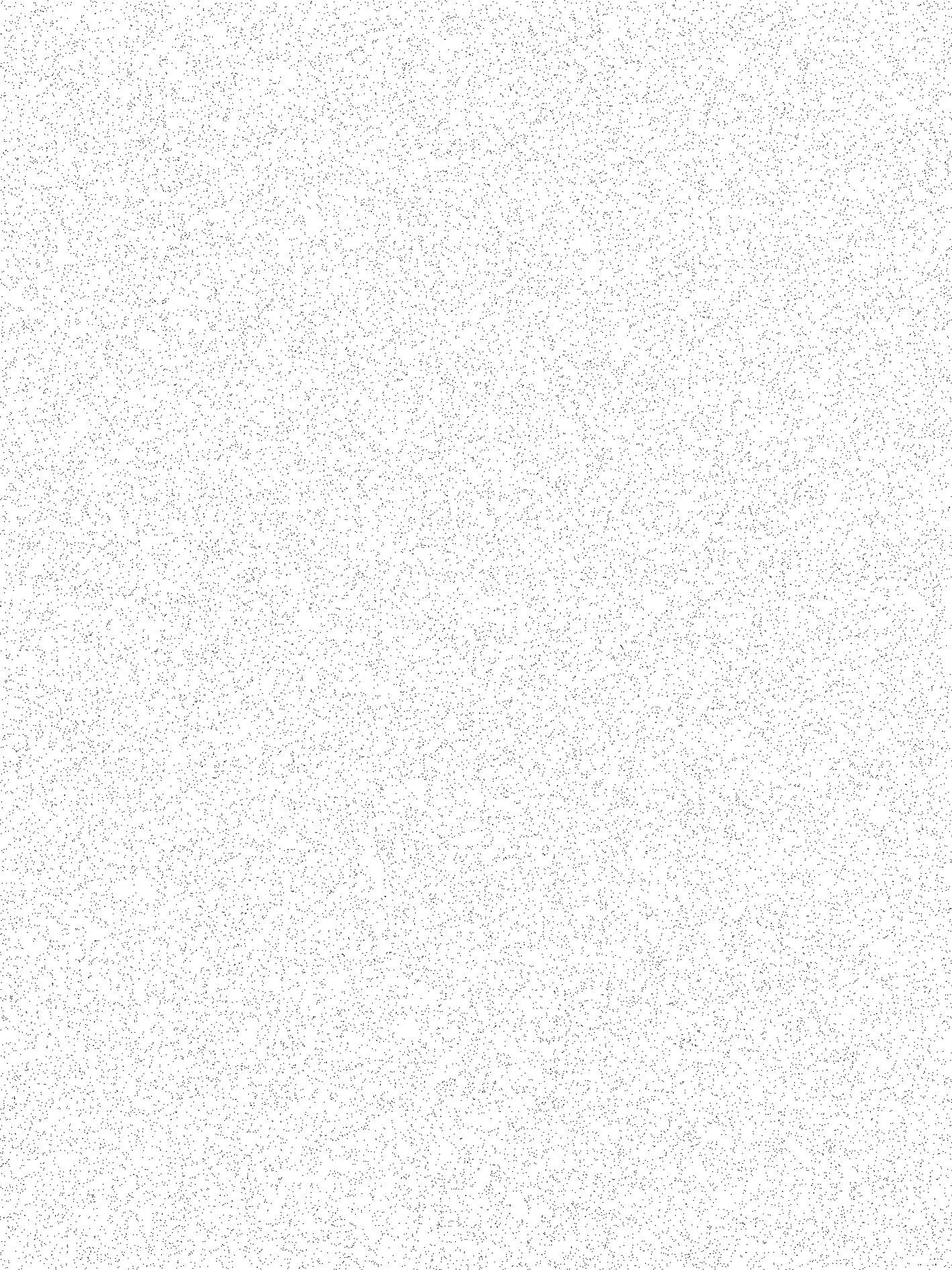}
\pgfsetupimageaspattern[width=4cm]{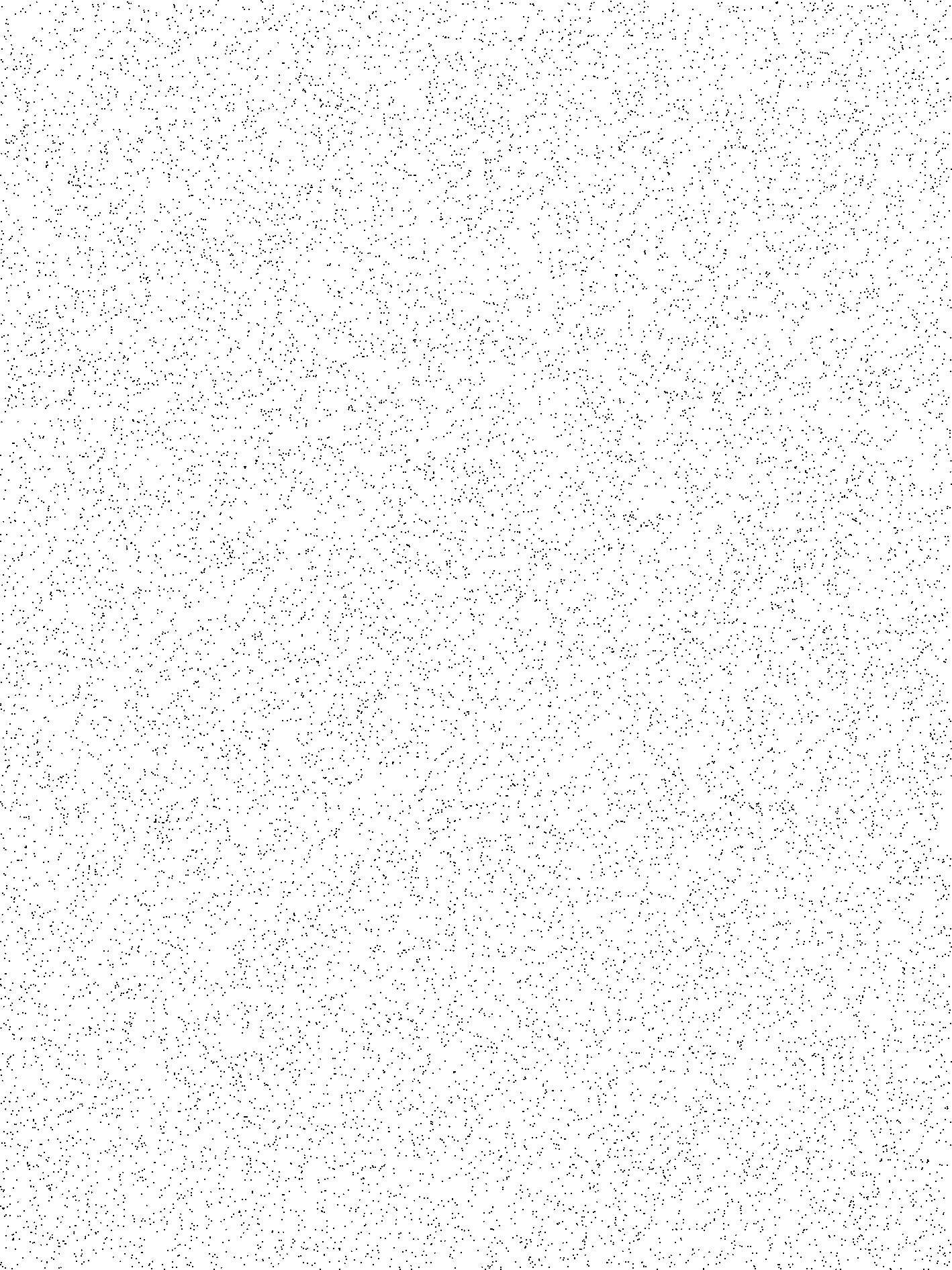}{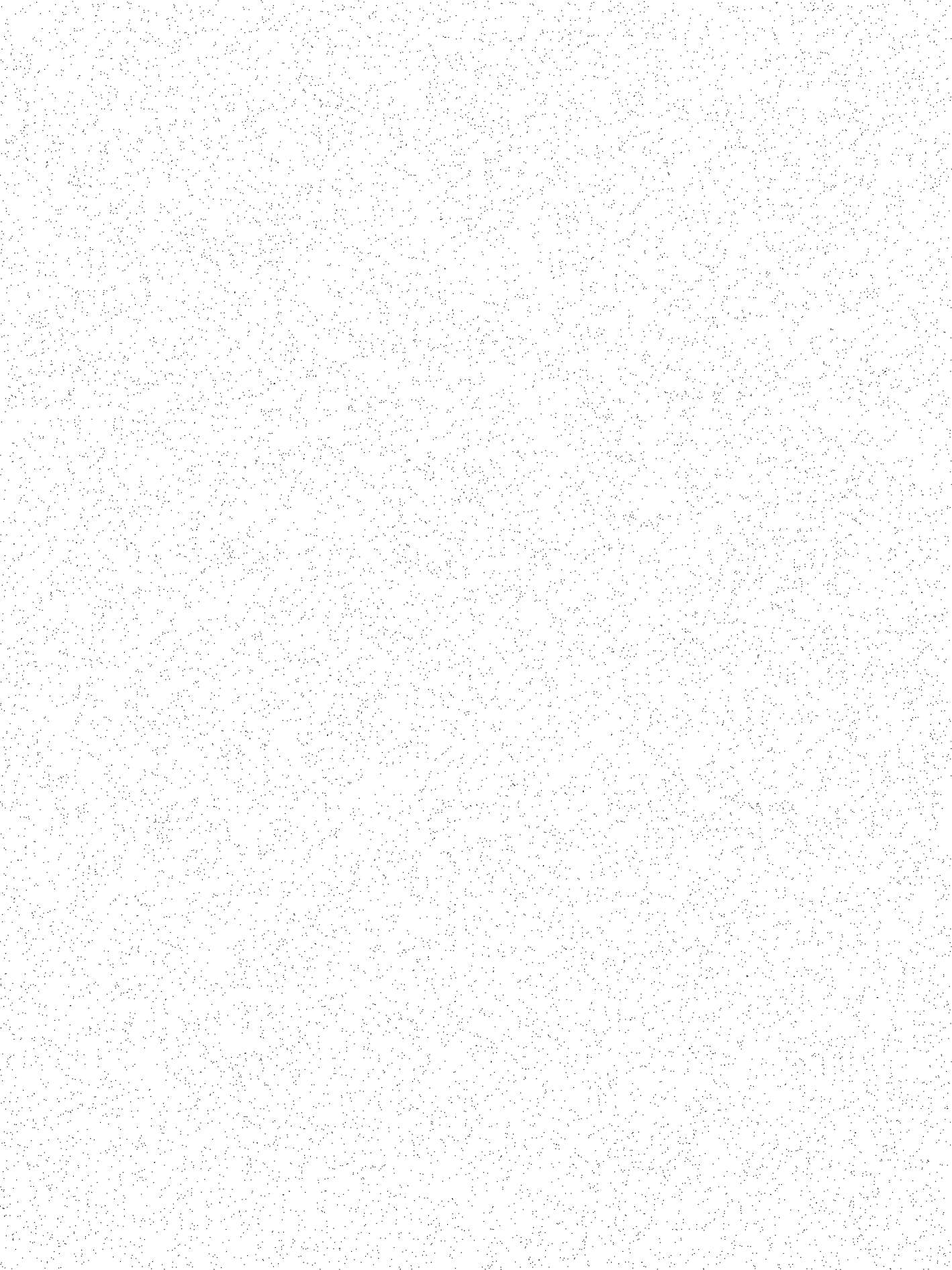}
\pgfsetupimageaspattern[width=4cm]{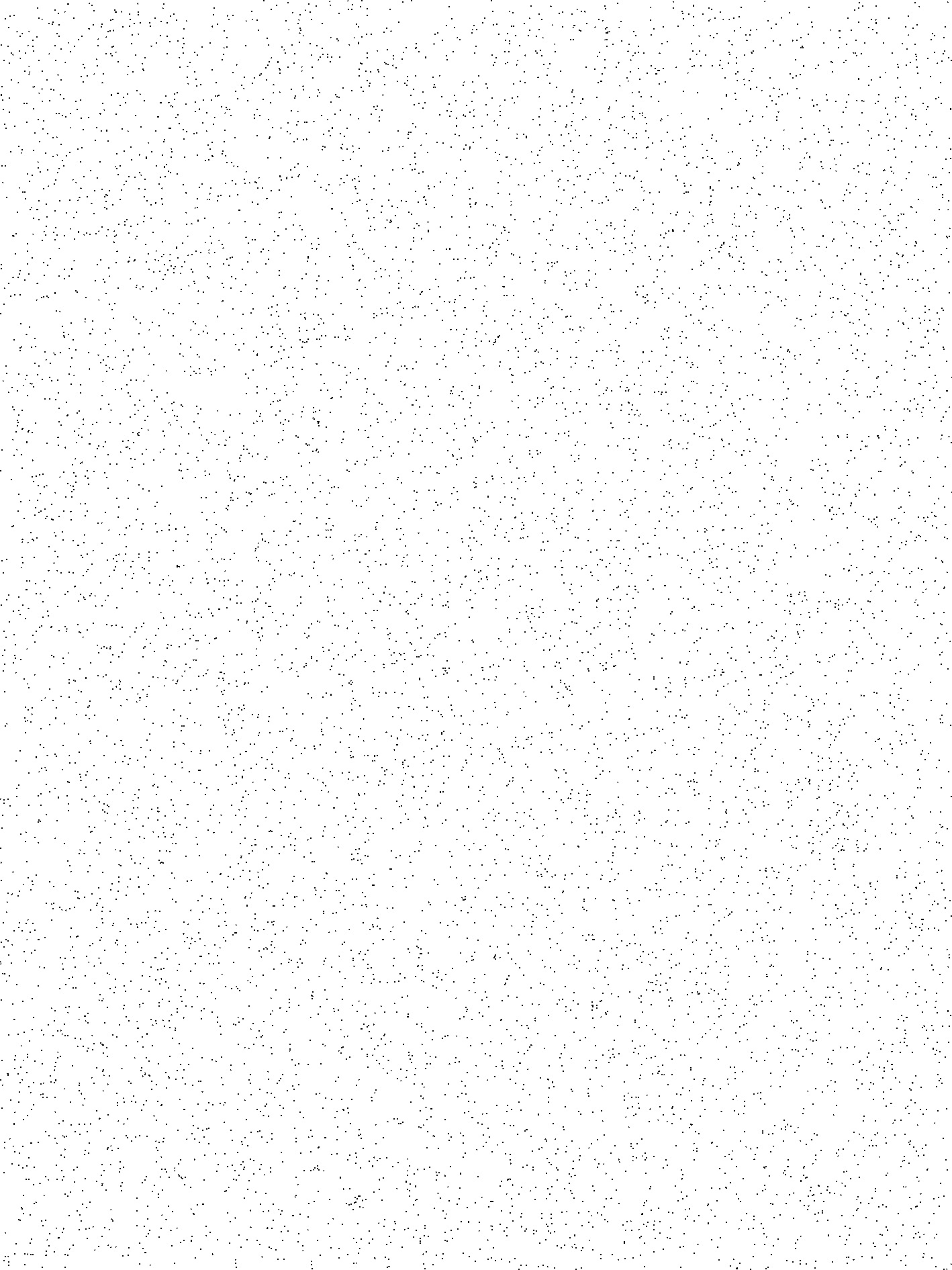}{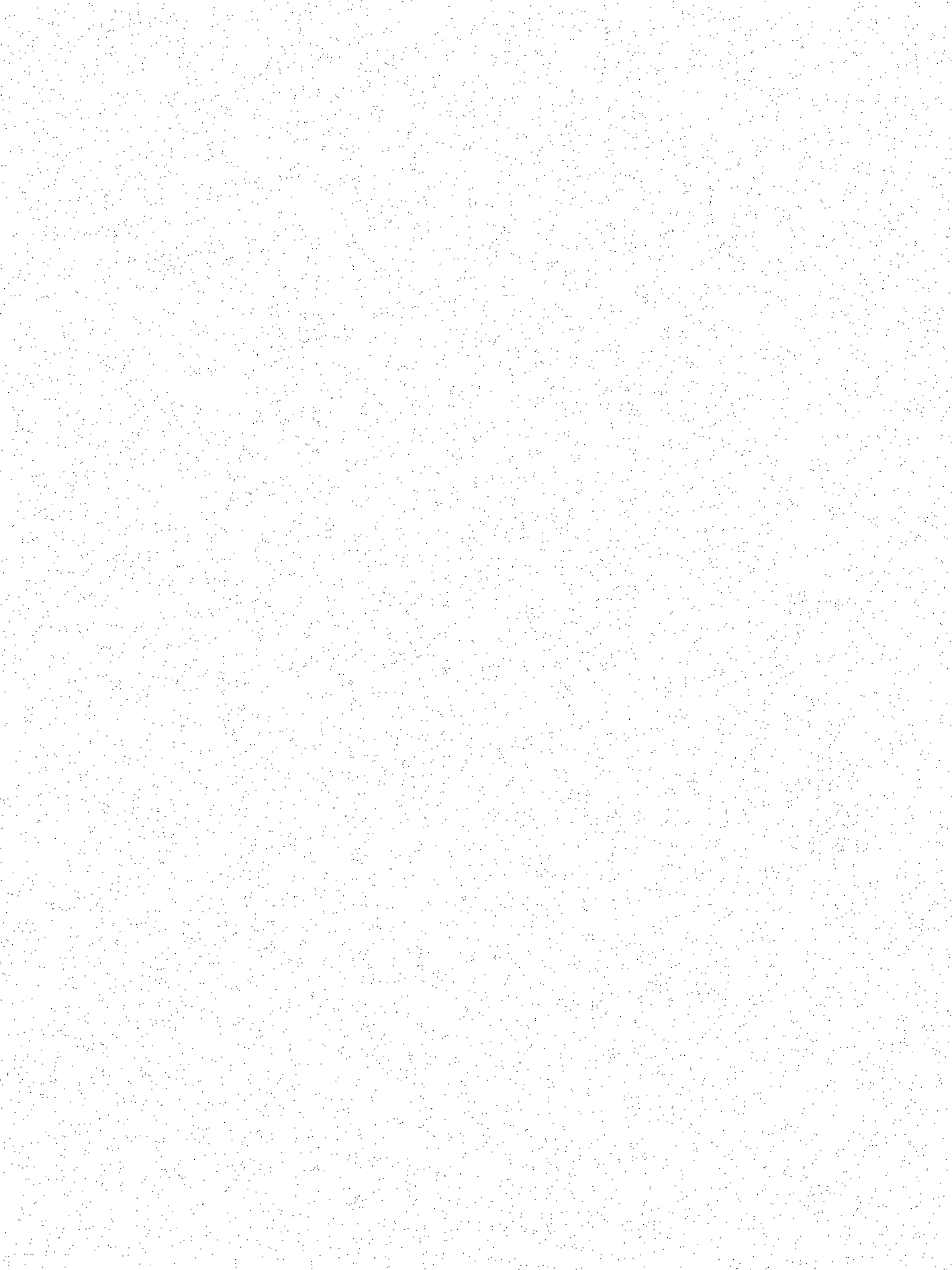}
\pgfsetupimageaspattern[width=4cm]{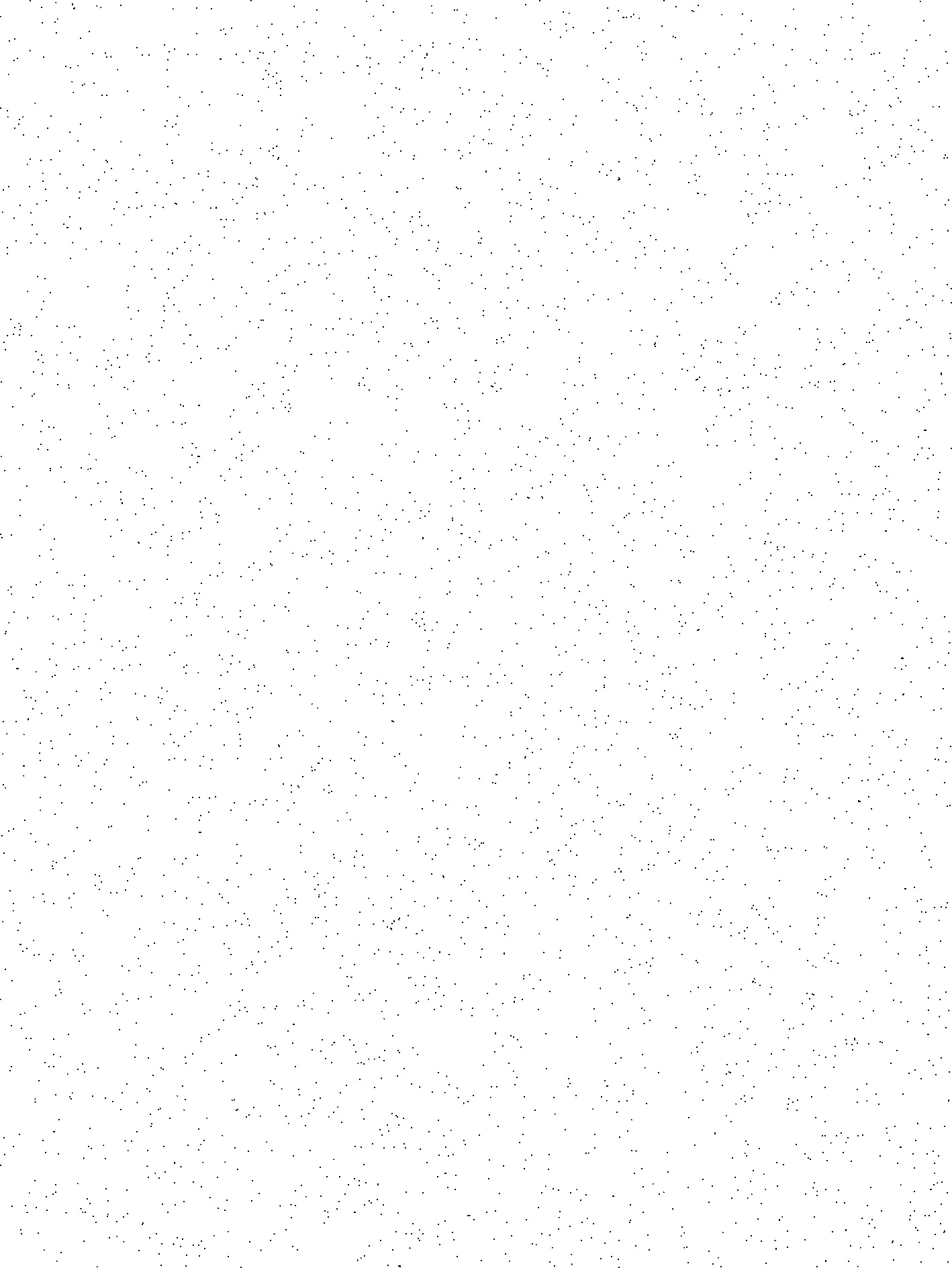}{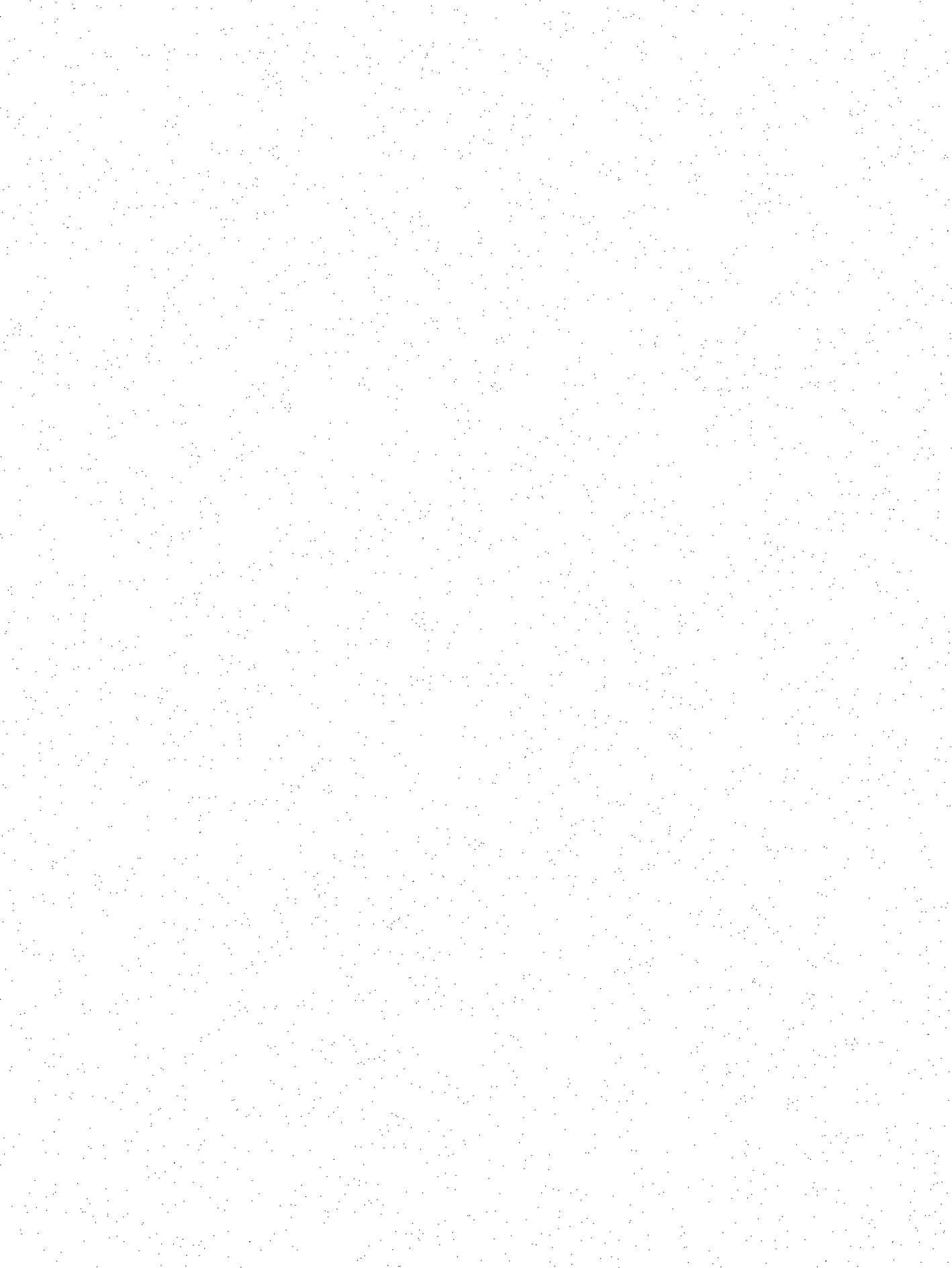}

%% file: sec/abstract.tex
\begin{abstract}
%
Bimanual manipulation is crucial in robotics, enabling complex tasks in industrial automation and household services. However, it poses significant challenges due to the high-dimensional action space and intricate coordination requirements. While video prediction has been recently studied for representation learning and control, leveraging its ability to capture rich dynamic and behavioral information, its potential for enhancing bimanual coordination remains underexplored. To bridge this gap, we propose a unified diffusion-based framework for the joint optimization of video and action prediction. Specifically, we propose a multi-frame latent prediction strategy that encodes future states in a compressed latent space, preserving task-relevant features. Furthermore, we introduce a unidirectional attention mechanism where video prediction is conditioned on the action, while action prediction remains independent of video prediction. This design allows us to omit video prediction during inference, significantly enhancing efficiency. Experiments on two simulated benchmarks and a real-world setting demonstrate a significant improvement in the success rate over the strong baseline ACT using our method, achieving a \textbf{24.9\%} increase on ALOHA, an \textbf{11.1\%} increase on RoboTwin, and a \textbf{32.5\%} increase in real-world experiments. Our models and code are publicly available at \href{https://github.com/return-sleep/Diffusion_based_imaginative_Coordination}{\texttt{Diffusion\_based\_imaginative\_Coordination}}.
\end{abstract}

%% file: sec/introduction.tex
\section{Introduction}







Bimanual manipulation \cite{smith2012dual,krebs2022bimanual_survey} is essential for a wide range of applications, including industrial manufacturing \cite{buhl2019dual,hermann2011hardware}, surgical operations \cite{wu2019coordinated,xie2024flexible}, and service automation \cite{zhang2024arp}. Compared to single-arm robots, bimanual robots offer superior dexterity, efficiency, and versatility by enabling coordinated, simultaneous actions with two arms, making them well-suited for complex, heavy, or precision-oriented tasks. However, bimanual manipulation presents significantly greater challenges than single-arm manipulation. One of the main challenges is the well-known \textit{multimodal behavior}. This phenomenon is more pronounced in bimanual manipulation compared to unimanual manipulation. For example, there are multiple possible ways to grasp an object: it can be held with the left hand, the right hand, or both hands. Another challenge is the need for precise \textit{spatiotemporal coordination}. For instance, in a pouring water task, both arms must be spatially aligned and temporally synchronized to ensure successful execution.

\begin{figure}[t]
    \centering
    \input{tikz_figures/intro}
    \caption{\textbf{Performance overview on bimanual manipulation tasks.}  Our method demonstrates significant improvements over previous approaches on ALOHA~\cite{zhao2023aloha}, RoboTwin~\cite{mu2024robotwin}, and real-world evaluations.}
    \label{fig:motivation_overview}
    \vspace{-3mm}
\end{figure}


To address the challenge posed by multimodality, prior works~\cite{zhao2023aloha, liu2024rdt, li2024planning} have explored the use of generative models, such as Conditional Variational Autoencoder (CVAE) ~\cite{sohn2015learning} and diffusion models~\cite{ddpm, nichol2021improved}, to capture the multimodal nature of bimanual manipulation. Regarding the challenge of coordination, some studies have introduced inductive biases to constrain functional roles \cite{grannen2023stabilize, liu2024voxactb}, regulate task execution \cite{mirrazavi2016coordinated, amadio2019exploiting}, or employ auxiliary modules to dynamically schedule actions for both arms \cite{zhao2023dualafford, lu2024anybimanual}. While these approaches achieve reasonable performance, they often lack flexibility, are designed for specific tasks, or rely on pretrained skill primitives.

In this paper, we aim to develop a bimanual manipulation model without relying on intricate, task-specific designs, drawing inspiration from how humans coordinate their two hands. Humans naturally imagine future scenarios, enabling them to plan actions with greater coordination and precision. Motivated by this, we explore an implicit coordination method that leverages future imagination as a consensus \cite{ren2008distributed} carrier for visuomotor policy. Our key insight is that conventional bimanual strategies, which rely solely on shared observations for coordination, are inherently limited by their lack of foresight into future interactions. Predicting future states allows both arms to anticipate interactions, align their objectives, and reach consensus.
The future visual states prediction is actually the video prediction problem~\cite{hafner2019dreamer}. Although video prediction-based action models has been widely studied in robot learning~\cite{wu2023gr-1,li2025gr_mg,hafner2019dreamer,unipi}, they are mainly limited to unimanual manipulation tasks, leaving video prediction for bimanual manipulation largely unexplored. Besides, the video prediction based models usually struggles with the issue of computation effecincy.

To address these challenges for bimanual manipulation, we propose a unified diffusion-based framework for jointly optimizing video and action prediction. This framework leverages diffusion models to enhance multimodal modeling, while video prediction contributes to improved bimanual coordination. To maximize the utility of video prediction, we enforce temporal consistency by predicting consecutive future frames and incorporating structured trajectory information. To enhance efficiency within the diffusion framework, we propose predicting future states using video tokens compressed by a video tokenizer. Compared to directly predicting frame pixels or image tokens, video tokens significantly reduce redundancy while preserving sufficient information for accurate video reconstruction. Additionally, to accelerate inference and meet the demands of high-frequency tasks, we introduce a unidirectional action-conditioned attention mechanism. During training, only video tokens are conditioned on action tokens, while action prediction remains independent of video tokens. Consequently, video prediction can be skipped during inference, significantly improving computational efficiency.


Experimental results demonstrate that our method consistently outperforms baseline models \cite{zhao2023aloha,chi2023diffusion,ze20243d_dp,li2025gr_mg} across two simulated benchmarks and a real-world setting. Specifically, on the ALOHA benchmark \cite{zhao2023aloha} for fine-grained bimanual manipulation, our method achieves a 24.9\% improvement in performance. On the RoboTwin benchmark \cite{mu2024robotwin}, which comprises 16 diverse tasks, our approach surpasses the previous state-of-the-art by 5.2\% in success rate. Furthermore, in real-world scenarios, our method maintains superior performance, achieving an average success rate of 60\% across 4 complex bimanual tasks. Our contributions can be summarized as follows:

\begin{itemize}

\item We introduce video prediction as a common imaginative interface, allowing both arms to anticipate future states and align their objectives. Unlike conventional approaches that rely solely on shared observations or predefined coordination rules, our method leverages the imagination of the future for implicit coordination.

\item  We propose a unified diffusion-based framework for jointly optimizing video and action prediction. We propose a multi-frame latent prediction task within this framework to preserve features relevant to action prediction. A unidirectional action-conditioned attention mechanism further integrates predictive information into action reasoning, ensuring robust performance and efficient inference.

\item  We evaluate our method on two simulated benchmarks and a real-world setting, achieving consistent improvements over state-of-the-art methods. Ablation studies further demonstrate the contribution of video prediction to bimanual manipulation and the effectiveness of the proposed modules.

\end{itemize}

%% file: tikz_figures/intro.tex
\begin{tikzpicture}
    

\node[draw, dashed, opacity=0.5, rounded corners, minimum width=\linewidth, minimum height=0.47\linewidth, anchor=north, yshift=-3pt]  (perf_rect) {}; 

\definecolor{darkseagreen11417699}{RGB}{114,176,99}
\definecolor{darkslateblue7495126}{RGB}{74,95,126}
\definecolor{silver184219179}{RGB}{184,219,179}

\begin{groupplot}[group style={
group size=3 by 1,
horizontal sep=4pt, 
},
axis lines=left,
axis line style={-},
width=\linewidth, 
ytick=\empty, ytick style={draw=none},
xtick=\empty, xtick style={draw=none},
]
\nextgroupplot[
tick pos=left,
xlabel={ALOHA~\cite{zhao2023aloha}},
xmin=0.0325, xmax=0.9675,
ylabel={Success Rate (\%)},
ymin=0, ymax=85,
at={($(perf_rect.south west)+(1.5em, 1.8em)$)},
anchor=south west,
width=1/2.05*\linewidth,
height=0.5*\linewidth,
ylabel style={yshift=-4pt}, 
legend columns=-1, 
legend style={fill=none,
      at={(0, 0)},
      draw=none,
      anchor=south,
      legend cell align=left,
      /tikz/every even column/.append style={column sep=8pt},
      legend to name = grouplegend,
      },
]
\draw[draw=none,fill=darkseagreen11417699,fill opacity=0.8,postaction={pattern=north west lines}] (axis cs:0.075,0) rectangle (axis cs:0.325,30.35);
\draw[draw=none,fill=silver184219179,fill opacity=0.8,postaction={pattern=horizontal lines}] (axis cs:0.375,0) rectangle (axis cs:0.625,47);
\draw[draw=none,fill=darkslateblue7495126,fill opacity=0.8,postaction={pattern=north east lines}] (axis cs:0.675,0) rectangle (axis cs:0.925,71.86);
\path [draw=black, dash pattern=on 3.7pt off 1.6pt, thick]
(axis cs:0.375,71.86)
--(axis cs:0.675,71.86);
\addlegendimage{ybar,area legend,draw=none,fill=darkseagreen11417699,fill opacity=0.8,postaction={pattern=north west lines}}
\addlegendentry{Diffusion Policy~\cite{chi2023diffusion}}
\addlegendimage{ybar,area legend,draw=none,fill=silver184219179,fill opacity=0.8,postaction={pattern=horizontal lines}}
\addlegendentry{ACT~\cite{zhao2023aloha}}
\addlegendimage{ybar,area legend,draw=none,fill=darkslateblue7495126,fill opacity=0.8,postaction={pattern=north east lines}}
\addlegendentry{Ours}

\draw (axis cs:0.5,71.86) ++(0pt,0pt) node[
  scale=0.7,
  anchor=south,
  text=black,
  rotate=0.0
]{\large\bfseries +24.9\%};
\draw[->,draw=black, thick, >=Stealth] (axis cs:0.5,47) -- (axis cs:0.5,71.86);

\nextgroupplot[
tick pos=left,
xlabel={RoboTwin~\cite{mu2024robotwin}},
xmin=0.0325, xmax=0.9675,
ymin=0, ymax=85,
width=1/2.05*\linewidth,
height=0.5*\linewidth,
]
\draw[draw=none,fill=darkseagreen11417699,fill opacity=0.8,postaction={pattern=north west lines}] (axis cs:0.075,0) rectangle (axis cs:0.325,27.77);
\draw[draw=none,fill=silver184219179,fill opacity=0.8,postaction={pattern=horizontal lines}] (axis cs:0.375,0) rectangle (axis cs:0.625,44.88);
\draw[draw=none,fill=darkslateblue7495126,fill opacity=0.8,postaction={pattern=north east lines}] (axis cs:0.675,0) rectangle (axis cs:0.925,55.98);
\path [draw=black, dash pattern=on 3.7pt off 1.6pt, thick]
(axis cs:0.375,55.98)
--(axis cs:0.675,55.98);

\draw (axis cs:0.5,55.98) ++(0pt,0pt) node[
  scale=0.7,
  anchor=south,
  text=black,
  rotate=0.0
]{\large\bfseries +11.1\%};
\draw[->,draw=black, thick, >=Stealth] (axis cs:0.5,44.88) -- (axis cs:0.5,55.98);

\nextgroupplot[
tick pos=left,
xlabel={Real-world},
xmin=0.0325, xmax=0.9675,
ymin=0, ymax=85,
width=1/2.05*\linewidth,
 height=0.5*\linewidth,
]
\draw[draw=none,fill=darkseagreen11417699,fill opacity=0.8,postaction={pattern=north west lines}] (axis cs:0.075,0) rectangle (axis cs:0.325,7.5);
\draw[draw=none,fill=silver184219179,fill opacity=0.8,postaction={pattern=horizontal lines}] (axis cs:0.375,0) rectangle (axis cs:0.625,27.5);
\draw[draw=none,fill=darkslateblue7495126,fill opacity=0.8,postaction={pattern=north east lines}] (axis cs:0.675,0) rectangle (axis cs:0.925,60);
\path [draw=black, dash pattern=on 3.7pt off 1.6pt, thick]
(axis cs:0.375,60)
--(axis cs:0.675,60);

\draw (axis cs:0.5,60) ++(0pt,0pt) node[
  scale=0.7,
  anchor=south,
  text=black,
  rotate=0.0
]{\large\bfseries +32.5\%};
\draw[->,draw=black, thick, >=Stealth] (axis cs:0.5,27.5) -- (axis cs:0.5,60);
\end{groupplot}
\node[anchor=north] at (perf_rect.north) {\ref*{grouplegend}};
\end{tikzpicture}

%% file: sec/related_work.tex
\section{Related Work}

\textbf{Learning-based bimanual manipulation.} Bimanual manipulation enables robots to tackle complex real-world tasks, such as household activities \cite{zhang2024empowering,varley2024embodied,gao2024bi,oh2024bimanual}. However, their high-dimensional action space poses significant challenges for complex behavior modeling and spatiotemporal coordination \cite{grotz2024peract2}. Early studies employed skill primitives to shrink the action space \cite{chitnis2020efficient,franzese2023interactive}. Recently, diffusion models have shown promise in bimanual policy learning, effectively capturing multimodal behaviors \cite{li2024planning, liu2024rdt, black2024pi_0,yu2024bikc,alohaunleashed}. To explicitly modeling coordination, some approaches introduce inductive biases, leveraging task symmetry assumptions \cite{amadio2019exploiting,li2023efficient} or leader-follower frameworks \cite{liu2022robot}. Additionally, predefined functional roles have been explored, where one arm performs manipulation while the other provides stabilization \cite{grannen2023stabilize,liu2024voxactb}.  However, such methods often lack flexibility and struggle to adapt to a variety of tasks. Recent studies have introduced auxiliary modules \cite{lu2024anybimanual,zhao2023dualafford} or large language models (LLMs) \cite{chu2024large,gao2024dag} as task schedulers, dynamically assigning tasks to each arm based on task demands to improve coordination. In contrast, our work jointly learns action fitting and video prediction to implicitly encourage coordination between two arms.

\noindent\textbf{Video prediction for robotics.} Videos capture rich information about dynamics and behavior, offering valuable opportunities for enhancing robot learning. Similar to next-token prediction in large language model pre-training, some works have leveraged next-frame prediction for robot manipulation tasks~\cite{wu2023gr-1,LAPA,MPI,unipi,AVDC}. One direction treats it as a pre-training or joint training task~\cite{wu2023gr-1,MPI,LAPA,VidMan,guo2024prediction}. In this line of work, the features are shared between video and action prediction tasks. The advantage of pre-training is the ability to transfer knowledge from human video datasets, such as Ego4D~\cite{grauman2022ego4d}, to manipulation tasks. For example, GR1~\cite{wu2023gr-1} first pre-trains a model on the Ego4D~\cite{grauman2022ego4d} dataset using a video prediction task. The model is then fine-tuned on a robot dataset with both video and action prediction tasks. Unlike GR1~\cite{wu2023gr-1}, which directly predicts the next RGB frame, our approach predicts latent tokens instead. Another line of works uses future frame prediction as a video plan, followed by an inverse dynamics model conditioned on this plan to generate actions~\cite{unipi,AVDC,SuSiE,SEER}. Compared to this line of work, our model predicts future frames and actions in parallel with a unidirectional attention mechanism, enabling \textit{significantly faster inference} by predicting only actions. Unlike most prior works, which lack evaluation on bimanual manipulation, our model is specifically designed for it and proves effective in both simulated and real-world experiments.

%% file: sec/method.tex
\section{Method}
\label{sec:Method}

\begin{figure*}[ht]
    \centering 
    \input{tikz_figures/overview}
    \caption{\textbf{Model Overview}. We formulate bimanual manipulation as a conditional generation problem and propose a unified transformer-based diffusion model for action prediction and video forecasting, optimizing the perception-prediction-control procedure in an end-to-end manner. Taking a single-view image and the robot's proprioceptive state as input, our method simultaneously predicts the future action sequence and corresponding frames. }
    \label{fig:model_overview} 
    \vspace{-3mm}
\end{figure*}


\subsection{Problem Formulation}

\textbf{Bimanual imitation learning.}
We aim to learn a bimanual policy $\pi$ from expert demonstrations, which consist of multiple robot trajectories:
\begin{equation}
    \tau = \{(o_1, p_1, a_1), \dots, (o_T, p_T, a_T)\},
\end{equation}
where $o_t$ represents the visual observation and $p_t$ denotes the robot proprioceptive state at timestep $t$. The action $a_t = (a^l_t, a^r_t)$ consists of the joint positions of the left and right arms, including the gripper states. Given the current observation as input, the policy is trained to predict the future action sequence in the subsequent $N$ timesteps:
\begin{equation}
    \pi(o_t, p_t) \rightarrow a_{t:t+N}.
\end{equation}

\noindent\textbf{Bimanual manipulation with video prediction.}
Due to the inherent connection between video prediction and action learning, we place both tasks under a unified conditional generative framework. Specifically, given the current observation $(o_t, p_t)$, the model jointly predicts both the future action sequence and the future visual observations:
\begin{equation}
    \pi(o_t, p_t) \rightarrow (a_{t:t+N}, o_{t+1:t+N+1}).
\end{equation}

 By jointly modeling video prediction and action learning, we construct a shared future imagination space, where visual perception, environment dynamics modeling, and action control are co-optimized.

\subsection{Model Architecture}
An overview of our model architecture is shown in~\cref{fig:model_overview}. We formulate bimanual manipulation as a conditional generation problem and propose a unified transformer-based diffusion model for action prediction and video forecasting.  First, given current observation, which includes the robot's joint states and visual inputs, the model encodes this information into an observation embedding. Second, this embedding serves as guidance for the denoising decoder, which iteratively refines noisy latent variables. Third, the model generates both future action and video representations.

\noindent\textbf{Observation tokenization and embedding.} The observation consists of both visual information and robot proprioceptive state, which are separately processed and tokenized.  Single-view RGB image is passed through a ResNet-18-based 2D image encoder to extract spatial features, producing a 2D feature map $F \in \mathbb{R} ^{C \times H\times W}$. The feature map is then flattened and projected into $H\times W$ visual tokens, each represented as a 
$d$-dimensional vector. A 2D positional encoding is added to each visual token to retain spatial structure. Similarly, the robot's proprioceptive state is mapped into the same latent space via a linear layer. The final observation representation is obtained by concatenating the visual tokens with the proprioceptive embedding, resulting in $H\times W+1 $ tokens. These tokens are processed by a transformer encoder composed of multiple stacked layers to construct the observation embedding $e$, which serves as the conditioning input for the diffusion model.

\noindent\textbf{Future frame representation.} To enable efficient video prediction, we compress future frames into a compact tokenized representation. Specifically, we perform uniform temporal sampling to ensure sufficient visual variation across frames and then encode the sampled frames using a pretrained Cosmos video tokenizer~\cite{agarwal2025cosmos}. Unlike image tokenizers used in previous works \cite{van2017vqvae,esser2021vqgan,rombach2022ldm}, video tokenizers can reduce temporal redundancy and significantly decrease computational complexity while retaining sufficient information for video reconstruction \cite{ma2024latte}. Similar to ViT \cite{dosovitskiy2020vit}, we apply a patchification process to the obtained latent features and divide them into non-overlapping patches, yielding final tokens. Each future token is further enriched with a spatio-temporal position embedding. This enriched token provides a structured and computationally efficient representation for future frames, allowing for more effective modeling of high-level temporal dynamics.

\noindent\textbf{Conditional diffusion modeling.} Given the previously obtained observation embedding $e$ and the diffusion step $k$, we employ a conditional diffusion model to iteratively refine noisy representations of future actions and visual tokens. At each diffusion step, noise is injected into the clean action sequence $a$ and future frame tokens $v$. These noisy representations are then projected into latent space via linear layers.  A shared transformer decoder processes these latent embeddings using cross-attention to extract information from the observation embedding and the diffusion step encoding. The final outputs are obtained through two separate prediction heads: (1) the action head, a linear layer that reconstructs the denoised action sequence, and (2) the prediction head, a multi-layer perceptron (MLP) that recovers the clean future visual tokens.

\begin{figure}
    \centering
    \input{tikz_figures/attention}
    \caption{\textbf{Illustration of unidirectional attention mechanism.} Action tokens attend exclusively to themselves, while future frame tokens attend to both themselves and historical action tokens. As for decoupled attention, actions and visual latent tokens attend only within their own modality.  It shows a scenario predicting four actions and two frames.}
    \label{fig:attention_mechanism}
    \vspace{-3mm}
\end{figure}
\noindent\textbf{Unidirectional attention mechanism.} Inspired by action-conditioned prediction \cite{oh2015action,nunes2020action,hafner2019dreamer}, we design a unidirectional attention mechanism within self-attention layer of  cross-attention module in the denoising decoder to regulate information flow between action and future frame tokens, shown in \cref{fig:attention_mechanism}.  Action tokens attend exclusively to themselves, while future frame tokens attend to both themselves and historical action tokens, enabling the model to capture the temporal dependencies between executed actions and their corresponding visual outcomes \footnote{In the unidirectional attention, we define prior\-action tokens as those preceding the predicted frames, and vice versa.}. This design offers two key advantages. First, during training, allowing future frame tokens to attend to past action tokens enables the model to capture the causal relationship between actions and future visual states, facilitating the learning of environment dynamics. Second, during inference, action generation is independent of future visual tokens, eliminating the need for video prediction. This reduces computational overhead, lowers inference latency, and improves control frequency, which is crucial for real-time robotic execution.

\subsection{Training and Inference}

\noindent\textbf{Training objective.}
We randomly sample a timestep $t$ and a diffusion step $k$, and add noise $ \epsilon = ( \epsilon^{a}, \epsilon^{v})$ into the clean data point consisting of an action chunk and future visual tokens. The model is trained to predict this added noise, allowing it to iteratively recover the clean action and visual tokens. We employ a \( \ell_1 \) loss for action reconstruction and a mean squared error (MSE) loss for visual token prediction. The overall training objective is formulated as:

\begin{equation*}
\mathcal{L} =\|\epsilon^{a}_\theta (o, p, k, a^k, v^k) - \hat{\epsilon}^{a}\|_1 
+w \cdot \|\epsilon^{v}_\theta (o, p, k, a^k, v^k) - \hat{\epsilon}^{v}\|_2^2.
\end{equation*}
The noisy action \( a^k \) and noisy visual tokens \( v^k \) are computed using a predefined noise scheduler, which determines the level of noise at each step. $w$ is a weighting coefficient that controls the relative importance of the video prediction. Please refer to the appendix for implementation details.

\noindent\textbf{Inference and execution.} During inference, video prediction is omitted, and only action generation is performed to reduce latency and improve control frequency. This is enabled by the unidirectional attention mechanism, where action tokens do not receive information from future visual tokens. Given an initial action sequence sampled from Gaussian noise,  $a^K \sim \mathcal{N}(0, I) $, the model iteratively refines the action tokens through the reverse diffusion process. The action chunk is executed in an open-loop manner until completion, after which the next action chunk is generated.


%% file: tikz_figures/overview.tex
\definecolor{ffnn_color}{RGB}{235,182,120}
\begin{tikzpicture}[
coder/.style={draw, 
    very thick, 
    rounded corners,
    minimum width=1/7*\linewidth,
    minimum height=1.618/7*\linewidth,
    fill=cyan!20,
    align=center
},
ffnn/.style={draw,
    rotate=270,
    minimum width=0.7/7*\linewidth, 
    minimum height=16pt,
    thick,
    fill=ffnn_color,
    align=center
},
photo/.style={
    inner sep=0pt, 
},
arrow/.style={->, thick, >={Stealth[length=2mm]}},
small_arrow/.style={
    arrow,
    very thin, 
    >={Stealth[length=1.5mm]},
},
token/.style={draw,
    minimum size=(2*0.7/7*\linewidth + 14pt-10mm)/6,
    fill=gray!20,
},
noised_token/.style args={#1}{
    token, 
    image as pattern={name=Rnd-Patt-#1},
},
v_token/.style={
    token, 
    fill=magenta!40,
},
noised_v_token/.style={
    v_token, 
    image as pattern={name=Rnd-Patt-250},
},
a_token/.style={
    token, 
    fill=olive!50,
},
noised_a_token/.style={
    a_token,
    image as pattern={name=Rnd-Patt-1000},
},
]

\node[ffnn] (cnn) {CNN};

\node[ffnn, anchor=west] at ($(cnn.east)-(0, 14pt)$) (mlp) {MLP};

\node[photo, anchor=east, xshift=-5mm, label=below:Camera] at (cnn.south) (camera) {
\includegraphics[width=2.2cm]{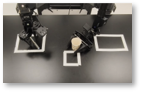}
};

\node[photo, anchor=east, xshift=-5mm, label=below:Prop. states] at (mlp.south)  (prop_state) {
\includegraphics[width=2.2cm]{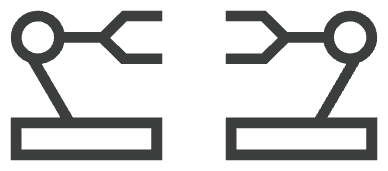}};

\draw[arrow] (camera) -- (cnn);
\draw[arrow] (prop_state) -- (mlp);

\pgfmathsetlengthmacro{\tokenSize}{(2*0.7/7*\linewidth + 14pt-10mm)/6};

\foreach \i in {0,...,5} {
    \ifthenelse{\i = 3}{
    \node[rotate=270,anchor=south west, yshift=8mm, inner sep=0, align=center] at ($(cnn.north west)+(0,-\i*\tokenSize - \i*2mm)$) (ct\i) {$\cdots$};
    }{
    \ifthenelse{\i = 0}{
    \node[token, anchor=north west, xshift=7mm] at ($(cnn.north west)+(0,0)$) (ct\i) {};
    }{
    \node[token, anchor=north west, xshift=7mm] at ($(cnn.north west)+(0,-\i*\tokenSize - \i*2mm)$) (ct\i) {};
    }}
}

\node[fit={(ct0)(ct4)},draw, thick, rounded corners] (cnn_out) {}; 
\coordinate (cnn_out_mid) at ($(cnn.north)!0.3!(cnn.north-|cnn_out.west)$);
\coordinate (mlp_out_mid) at ($(mlp.north)!0.3!(mlp.north-|ct5.west)$);

\draw[arrow, rounded corners] (mlp.north) -- (mlp_out_mid)|-(ct5.west);
\draw[arrow, rounded corners] (cnn.north) -- (cnn_out_mid) |- (cnn_out.west);

\coordinate (enc_input_mid) at ($(ct0.east)!0.5!(ct5.east)$);

\node[coder, anchor=west] at ($(enc_input_mid) + (3.5mm, 0)$) (encoder) {Transformer\\encoder}; 

\foreach \i in {0,...,5}{
    \ifthenelse{\i = 3}{
    }{
        \draw[small_arrow] (ct\i) -- (ct\i -| encoder.west);
    }
}

\foreach \i in {0,...,5} {
    \ifthenelse{\i = 3}{
    \node[rotate=270, anchor=south west, yshift=5mm, inner sep=0, align=center] at ($(encoder.north east)+(0,-\i*\tokenSize - \i*2mm)$) (e_ct\i) {$\cdots$};
    }{
    \ifthenelse{\i = 0}{
    \node[token, anchor=north west, xshift=3.5mm] at ($(encoder.north east)+(0,0)$) (e_ct\i) {};
    }{
    \node[token, anchor=north west, xshift=3.5mm] at ($(encoder.north east)+(0,-\i*\tokenSize - \i*2mm)$) (e_ct\i) {};
    }
    
    \draw[small_arrow] (e_ct\i -| encoder.east) -- (e_ct\i.west);
    }
}

\coordinate (enc_emb_mid) at ($(encoder.east)!0.7!(e_ct1.west)$); 

\node[fit={(e_ct0)(e_ct5)(enc_emb_mid)},draw, thick, rounded corners] (obs_embds) {};

\foreach \i in {0,...,2} {
    \ifthenelse{\i = 1}{
    \node[rotate=270, anchor=south west, yshift=14.5mm, inner sep=0, align=center] at ($(e_ct0.north east)+(0,-\i*\tokenSize - \i*2mm)$) (noised_tok\i) {$\cdots$};
    }{
    \node[noised_v_token, anchor=north west, xshift=18mm] at ($(e_ct0.north west)+(0,-\i*\tokenSize - \i*2mm)$) (noised_tok\i) {};
    }
}

\foreach \i in {3,...,5} {
    \ifthenelse{\i = 4}{
    \node[rotate=270, anchor=south west, yshift=14.5mm, inner sep=0, align=center] at ($(e_ct0.north east)+(0,-\i*\tokenSize - \i*2mm)$) (noised_tok\i) {$\cdots$};
    }{
    \node[noised_a_token, anchor=north west, xshift=18mm] at ($(e_ct0.north west)+(0,-\i*\tokenSize - \i*2mm)$) (noised_tok\i) {};
    }
}

\node[fit={(noised_tok0)(noised_tok2)}, thick, draw, rounded corners, inner sep=2pt] (noised_v_tokens) {};
\node[fit={(noised_tok3)(noised_tok5)}, thick, draw, rounded corners, inner sep=2pt] (noised_a_tokens) {};

\node[rotate=-90, anchor=north, align=center, font=\tiny\linespread{0.8}\selectfont] at (noised_v_tokens.west) (noised_v_tokens_label) {\small Noised video\\\small tokens};
\node[rotate=-90, anchor=north, align=center,font=\tiny\linespread{0.8}\selectfont] at (noised_a_tokens.west) (noised_a_tokens_label) {\small Noised action\\\small tokens};

\coordinate (noised_tokens_mid) at ($(noised_tok0)!.5!(noised_tok5)$); 

\node[coder, anchor=west, xshift=5.5mm] at (noised_tokens_mid.east) (decoder) {Joint\\denoising\\decoder};

\foreach \i in {0,...,5} {
    \ifthenelse{\i =1 \OR \i = 4}{}{
    \draw[small_arrow] (noised_tok\i) -- (noised_tok\i -| decoder.west);
    }
}

\foreach \i in {0,...,2} {
    \ifthenelse{\i = 1}{
    \node[rotate=270, anchor=south west, yshift=5mm, inner sep=0, align=center] at ($(decoder.north east)+(0,-\i*\tokenSize - \i*2mm)$) (d_ct\i) {$\cdots$};
    }{
    \ifthenelse{\i = 0}{
    \node[v_token, anchor=north west, xshift=3.5mm] at ($(decoder.north east)+(0,0)$) (d_ct\i) {};
    }{
    \node[v_token, anchor=north west, xshift=3.5mm] at ($(decoder.north east)+(0,-\i*\tokenSize - \i*2mm)$) (d_ct\i) {};
    }
    
    \draw[small_arrow] (d_ct\i -| decoder.east) -- (d_ct\i.west);
    }
}

\foreach \i in {3,...,5} {
    \ifthenelse{\i = 4}{
    \node[rotate=270, anchor=south west, yshift=5mm, inner sep=0, align=center] at ($(decoder.north east)+(0,-\i*\tokenSize - \i*2mm)$) (d_ct\i) {$\cdots$};
    }{
    \ifthenelse{\i = 3}{
    \node[a_token, anchor=north west, xshift=3.5mm] at ($(decoder.north east)+(0,-\i*\tokenSize - \i*2mm)$) (d_ct\i) {};
    }{
    \node[a_token, anchor=north west, xshift=3.5mm] at ($(decoder.north east)+(0,-\i*\tokenSize - \i*2mm)$) (d_ct\i) {};
    }
    
    \draw[small_arrow] (d_ct\i -| decoder.east) -- (d_ct\i.west);
    }
}

\coordinate (decoder_v_output_mid) at ($(d_ct0)!0.6!(decoder.east)$);
\coordinate (decoder_a_output_mid) at ($(d_ct3)!0.6!(decoder.east)$);

\node[fit={(d_ct0)(d_ct2)(decoder_v_output_mid)}, thick, draw, rounded corners, inner sep=2pt] (v_tokens) {};
\node[fit={(d_ct3)(d_ct5)(decoder_a_output_mid)}, thick, draw, rounded corners, inner sep=2pt] (a_tokens) {};

\coordinate (decoder_north) at ($(decoder.north) + (0, 8pt)$); 

\draw[arrow, rounded corners] (obs_embds.north)|-(decoder_north)--(decoder.north);

\node[photo, anchor=west, xshift=8mm, label=below:Future frames] at (v_tokens.east) (future_frames) {
\includegraphics[width=2.2cm]{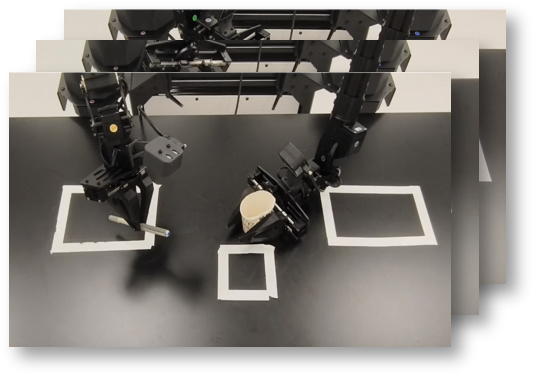}
};

\node[photo, anchor=west, xshift=8mm, label=below:Future actions] at (a_tokens.east)  (future_actions) {
\includegraphics[width=2.2cm]{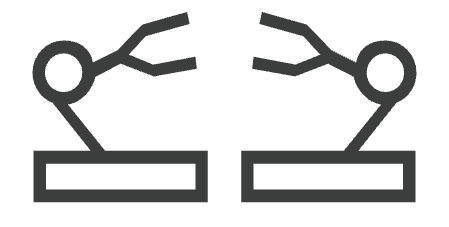}};

\draw[arrow] (v_tokens) -- (future_frames);
\draw[arrow] (a_tokens) -- (future_actions);

\node[anchor=south, yshift=0.5em] at ($(obs_embds.north)!.5!(decoder.north)$) (obv_embds_txt) {\small Observation embeddings};
\end{tikzpicture}

%% file: tikz_figures/attention.tex
\begin{tikzpicture}[font=\small]

\def\gridLen{1/22*\linewidth}

\foreach \x in {0,...,5} {
    \foreach \y in {0,...,5}{
    \node[draw=black, fill=red!20, minimum size=\gridLen] at (\x*\gridLen,\y*\gridLen) (full_\x_\y){};
    }
}

\draw[decorate, decoration={brace,amplitude=5pt}] (-\gridLen/2, -\gridLen/2) -- (-\gridLen/2, 3*\gridLen+\gridLen/2) node[midway, left, xshift=-\gridLen/2] {\rotatebox{90}{Action}};

\draw[decorate, decoration={brace,amplitude=5pt}] (0-\gridLen/2, 4*\gridLen-\gridLen/2) -- (-\gridLen/2, 5*\gridLen+\gridLen/2) node[midway, left, xshift=-\gridLen/2] {\rotatebox{90}{Latent}};

\draw[decorate, decoration={brace,amplitude=5pt}] (-\gridLen/2, 5*\gridLen+\gridLen/2) -- (2*\gridLen-\gridLen/2, 5*\gridLen+\gridLen/2) node[midway, above,yshift=4pt] {Latent};

\draw[decorate, decoration={brace,amplitude=5pt}] (2*\gridLen-\gridLen/2, 5*\gridLen+\gridLen/2) -- (5*\gridLen+\gridLen/2, 5*\gridLen+\gridLen/2) node[midway, above,yshift=4pt] {Action};

\coordinate (full_attetion_center) at ($(full_0_0)!.50!(full_5_5)$);

\node[yshift=-4.1*\gridLen, align=center,font=\small\linespread{0.8}\selectfont] at (full_attetion_center) (full_attetion_caption) {Full\\attention};

\def\deAttOffset{6.5\gridLen}
\foreach \x in {0,...,5} {
    \foreach \y in {0,...,5}{
    \ifthenelse{\x < 2 \AND \y > 3} {
    \node[draw=black, fill=red!20, minimum size=\gridLen] at (\x*\gridLen+\deAttOffset,\y*\gridLen) (de_\x_\y){};
    }{
    \ifthenelse{\x > 1 \AND \y < 4} {
    \node[draw=black, fill=red!20, minimum size=\gridLen] at (\x*\gridLen+\deAttOffset,\y*\gridLen) (de_\x_\y){};
    }{
    \node[draw=black, minimum size=\gridLen] at (\x*\gridLen+\deAttOffset,\y*\gridLen) (de_\x_\y){};
    }
    }
    }
}

\draw[decorate, decoration={brace,amplitude=5pt}] (-\gridLen/2+\deAttOffset, 5*\gridLen+\gridLen/2) -- (2*\gridLen-\gridLen/2+\deAttOffset, 5*\gridLen+\gridLen/2) node[midway, above,yshift=4pt] { Latent};

\draw[decorate, decoration={brace,amplitude=5pt}] (2*\gridLen-\gridLen/2+\deAttOffset, 5*\gridLen+\gridLen/2) -- (5*\gridLen+\gridLen/2+\deAttOffset, 5*\gridLen+\gridLen/2) node[midway, above,yshift=4pt] {Action};

\coordinate (de_attetion_center) at ($(de_0_0)!.50!(de_5_5)$);

\node[xshift=\deAttOffset, align=center,font=\small\linespread{0.8}\selectfont] at (full_attetion_caption) (de_attetion_caption) {Decoupled\\ attention};

\def\uniAttOffset{2*6.5\gridLen}
\foreach \x in {0,...,5} {
    \foreach \y in {0,...,5}{
    \ifthenelse{\x < 4 \AND \y > 3} {
    \node[draw=black, fill=red!20, minimum size=\gridLen] at (\x*\gridLen+\uniAttOffset,\y*\gridLen) (uni_\x_\y){};
    }{
    \ifthenelse{\x > 1 \AND \y < 4} {
    \node[draw=black, fill=red!20, minimum size=\gridLen] at (\x*\gridLen+\uniAttOffset,\y*\gridLen) (uni_\x_\y){};
    }{
    \node[draw=black, minimum size=\gridLen] at (\x*\gridLen+\uniAttOffset,\y*\gridLen) (uni_\x_\y){};
    }
    }
    }
}

\draw[decorate, decoration={brace,amplitude=5pt}] (-\gridLen/2+\uniAttOffset, 5*\gridLen+\gridLen/2) -- (2*\gridLen-\gridLen/2+\uniAttOffset, 5*\gridLen+\gridLen/2) node[midway, above,yshift=4pt] {\footnotesize Latent};

\draw[decorate, decoration={brace,amplitude=5pt}] (2*\gridLen-\gridLen/2+\uniAttOffset, 5*\gridLen+\gridLen/2) -- (3*\gridLen+\gridLen/2+\uniAttOffset, 5*\gridLen+\gridLen/2) node[midway, above,yshift=4pt,align=center, font=\scriptsize\linespread{0.8}\selectfont] {Prior-\\action};
\draw[decorate, decoration={brace,amplitude=5pt}] (4*\gridLen-\gridLen/2+\uniAttOffset, 5*\gridLen+\gridLen/2) -- (5*\gridLen+\gridLen/2+\uniAttOffset, 5*\gridLen+\gridLen/2) node[midway, above,yshift=4pt,align=center, font=\scriptsize\linespread{0.8}\selectfont] {Post-\\action};

\coordinate (uni_attetion_center) at ($(uni_0_0)!.50!(uni_5_5)$);

\node[xshift=\deAttOffset, align=center, font=\small\linespread{0.8}\selectfont] at (de_attetion_caption) () {Unidirectional\\ attention};

\end{tikzpicture}

%% file: sec/experiments.tex
\section{Experiments}
\label{sec:Experiments}

We comprehensively evaluate our method across multiple bimanual benchmarks, including the simulated ALOHA \cite{zhao2023aloha} and RoboTwin \cite{mu2024robotwin}, and further validate it in real-world environment. Our experiments aim to address  the following key questions:
\begin{enumerate}
\item Does our approach surpass existing methods for bimanual manipulation (see~\cref{subsec:Comparison with the State-of-the-Art Methods})?
\item What role does video prediction play in bimanual tasks, and how do design choices of prediction strategies, and attention mechanisms contribute to performance (see~\cref{subsec:Ablation Study})?
\item Can our model maintain its superior performance in real-world environments (see~\cref{subsec:Real-world Evaluation})?
\end{enumerate}

\subsection{Experiment Setup}


\noindent\textbf{Baselines.} We compare our method against several baselines to validate the effectiveness:

\begin{itemize}
    \item \textbf{ACT} \cite{zhao2023aloha}: a CVAE-based imitation learning method that utilizes action chunking to predict sequences of future actions and employs temporal ensemble to ensure smooth execution.
The model is inspired by DETR~\cite{detr}, which adopts an encoder-decoder architecture. Action queries are fed into a transformer decoder to predict the actions.

    \item \textbf{Diffusion Policy} \cite{chi2023diffusion}: a visuomotor policy learning framework that formulates action prediction as a conditional denoising diffusion process. 
     
    \item \textbf{3D Diffusion Policy} \cite{ze20243d_dp}: leverages a lightweight MLP encoder to process sparse point clouds and a conditional diffusion model to generate actions, enabling efficient and robust visuomotor policy learning.

    \item \textbf{RDT-1B} \cite{liu2024rdt}: a diffusion-based Transformer model for bimanual robotic manipulation, leveraging a unified action space and multi-modal inputs to achieve efficient few-shot learning across diverse tasks
   
    \item \textbf{GR-MG} \cite{li2025gr_mg}: an enhanced version of GR-1~\cite{wu2023gr-1}, retaining the same model architecture and next-frame prediction loss. It improves GR-1~\cite{wu2023gr-1} by incorporating a multi-modal, goal-conditioned generative policy that integrates language and goal images for robotic manipulation. This enables efficient learning from partially annotated data through progress-guided goal image generation.
\end{itemize}

\subsection{Comparison with State-of-the-Art Methods}
\label{subsec:Comparison with the State-of-the-Art Methods}

\begin{figure}
    \centering
    \includegraphics[width= 0.95\linewidth]{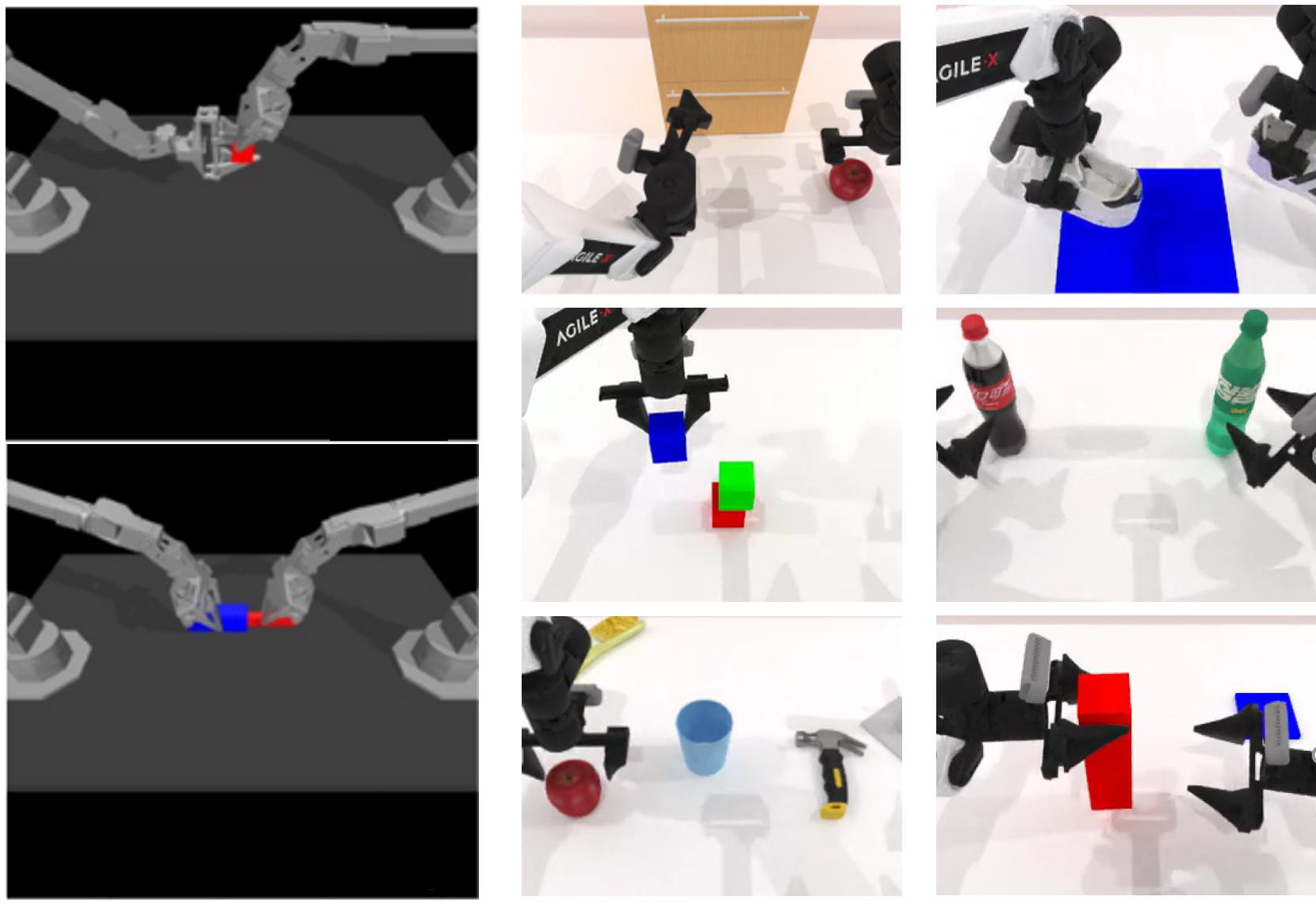}
    \caption{\textbf{Visualizations of bimanual tasks among two simulation benchmarks, including ALOHA~\cite{zhao2023aloha} and RoboTwin~\cite{mu2024robotwin}.}}
    \label{fig:via_sim_taskl}
    \vspace{-3mm}
\end{figure}

\begin{table}[t]
    \centering
    \resizebox{\linewidth}{!}{
    \begin{tabular}{@{}l c c c c c@{}}
        \toprule
         & \cellcolor{gray!20} Avg.$\uparrow$ &  \multicolumn{2}{c}{Transfer Cube} & \multicolumn{2}{c}{Insertion}  \\
        Method  & \cellcolor{gray!20} Success $(\%)$  & Scripted & Human & Scripted & Human \\
        \midrule
        ACT \cite{zhao2023aloha} & \cellcolor{gray!20}47 & 86 & 50 & 32 & 20  \\
        DP \cite{chi2023diffusion} & \cellcolor{gray!20}33  & 54 & 4 & 74 & 0\\

        Ours & \cellcolor{gray!20}\textbf{71.9} & \textbf{95.9} & \textbf{78.1} &\textbf{ 83.2} & \textbf{30.2}  \\
        \bottomrule
    \end{tabular}}
    \caption{\textbf{Evaluation on ALOHA benchmark.} We report the average success rates across 3 random seeds. }
    \label{tab:aloha_results}
    \vspace{-3mm}
\end{table}

\subsubsection{Results on ALOHA benchmark}

\noindent\textbf{ALOHA benchmark.} ALOHA benchmark \cite{zhao2023aloha} is built upon MuJoCo~\cite{todorov2012mujoco} and focuses on fine-grained bimanual manipulation.  ALOHA consists of two tasks: \textit{transfer cube} and \textit{bimanual insertion}. Demonstrations are collected by scripted policy or human teleoperation. Each task includes 50 demonstrations, spanning 400 steps (8 s).

\begin{table*}[ht]
    \centering
    \resizebox{0.98\linewidth}{!}{
    \begin{tabular}{@{}l l l l l  l l l l l @{}}
        \toprule
         & \cellcolor{gray!20} Avg.$\uparrow$ & Block & Block & Blocks  & Blocks & Bottle & Container & Diverse  & Dual Bottles   \\
        
        Method &\cellcolor{gray!20} Success $(\%)$  & Hammer  Beat & Handover & Stack (Easy) & Stack (Hard) & Adjust & Place & Bottles Pick & Pick (Easy)  \\
        \midrule
        DP3~\cite{ze20243d_dp} & \cellcolor{gray!20} 42.3 & 55.7\rbf{± 0.6} & 77.3 \rbf{± 11.6} & - & - & \textbf{73.3} \rbf{± 12.5} & \textbf{85.3} \rbf{± 3.2} & \textbf{37.0} \rbf{± 10.0} & 55.3 \rbf{± 11.5} \\
        DP~\cite{chi2023diffusion} & \cellcolor{gray!20} 27.3 & 0.0 \rbf{± 0.0} & 76.0 \rbf{± 16.1} & 8.0 \rbf{± 4.4} & 0.3 \rbf{± 0.6} & 35.7 \rbf{± 2.9} & 14.0 \rbf{± 6.9} & 12.0 \rbf{± 5.3} & 85.7 \rbf{± 6.7}  \\
        GR-MG \cite{li2025gr_mg} & \cellcolor{gray!20} 4.31 &  10.0 \rbf{± 12.1} & 0.0 \rbf{± 0.0} & 0.0 \rbf{± 0.0} & 0.0 \rbf{± 0.0} & 0.0 \rbf{± 0.0} &	0.0 \rbf{± 0.0} &2.7 \rbf{± 2.5}&30.3 \rbf{± 7.1} \\
        ACT \cite{zhao2023aloha} & \cellcolor{gray!20} 44.9 & 60.7 \rbf{± 18.6} &91.0 \rbf{± 3.0} & 54.0 \rbf{± 6.1} & 15.3 \rbf{± 6.5} & 51.7 \rbf{± 8.1} & 41.3 \rbf{± 5.7} & 13.7 \rbf{± 5.7} & \underline{89.7} \rbf{± 1.5}  \\
        
         RDT-1B \cite{liu2024rdt} & \cellcolor{gray!20} \underline{51.1} & \underline{94.3} \rbf{± 1.2} &  \underline{98.7} \rbf{± 0.6} & \underline{62.3} \rbf{± 4.9} & \underline{36.0} \rbf{± 1.0} & \underline{61.7} \rbf{± 8.4} & 52.7 \rbf{± 4.0} & 15.0 \rbf{± 0.0} &81.7 \rbf{± 2.9} \\
         
        Ours & \cellcolor{gray!20} \textbf{56.3} & \textbf{96.7} \rbf{± 2.5} & \textbf{100.0} \rbf{± 0.0} & \textbf{70.7} \rbf{± 1.5} & \textbf{37.7} \rbf{± 9.1} & 53.3 \rbf{± 7.8} & \underline{53.7} \rbf{± 2.5} & \underline{27.3} \rbf{± 1.5} & \textbf{94.7} \rbf{± 1.5} \\
        
        \midrule
          &  Dual Bottles 
         & Dual Shoes  & Empty Cup & Mug  Hanging   & Mug  Hanging & Pick Apple  & Put Apple & Shoe \\ 
        Method  &   Pick (Hard) 
         & Place & Place &(Easy) & (Hard) &  Messy &  Cabinet & Place  \\  
        \midrule
        DP3~\cite{ze20243d_dp} &  58.0 \rbf{± 3.0} & 12.0 \rbf{± 1.7} & \underline{61.7} \rbf{± 13.1} & \textbf{15.3} \rbf{± 4.0} & \textbf{15.3} \rbf{± 5.5} & 9.7 \rbf{± 2.1} & 66.3 \rbf{± 22.3} & 54.3 \rbf{± 0.6} \\ 
        DP~\cite{chi2023diffusion} & \underline{59.3} \rbf{± 5.5} & 3.0 \rbf{± 1.0} & \textbf{87.7} \rbf{± 0.6} & 0.0 \rbf{± 0.0} & 0.0 \rbf{± 0.0} & 29.3 \rbf{± 5.0} & 8.0 \rbf{± 12.2} & 25.3 \rbf{± 7.6} \\ 
        GR-MG \cite{li2025gr_mg} & 17.7 \rbf{± 7.7} & 0.0 \rbf{± 0.0} & 0.0 \rbf{± 0.0} &	0.0 \rbf{± 0.0} & 0.0 \rbf{± 0.0} & 8.0 \rbf{± 7.9}	& 0.0 \rbf{± 0.0}	& 0.3 \rbf{± 0.5} \\
        ACT \cite{zhao2023aloha} & 52.3 \rbf{± 2.3} & 18.7 \rbf{± 7.2} & 55.3 \rbf{± 7.2} & \underline{0.3} \rbf{± 0.6} & 2.7 \rbf{± 2.5} & 36.7 \rbf{± 12.6} & \underline{73.3} \rbf{± 10.3} & 61.3 \rbf{± 5.0}  \\ 
         RDT-1B \cite{liu2024rdt} & \textbf{60.0} \rbf{± 3.0} & \textbf{23.7} \rbf{± 3.5} & \underline{79.3} \rbf{± 3.1} & 0.0 \rbf{± 0.0} & 2.7 \rbf{± 0.6} & \underline{44.7} \rbf{± 7.1} & \textbf{31.3} \rbf{± 2.1} & \textbf{74.0} \rbf{± 1.0}  \\
         
        Ours & 55.3 \rbf{± 2.3} & \underline{20.3} \rbf{± 2.1} & 58.0 \rbf{± 6.1} & 0.0 \rbf{± 0.0} & \underline{3.0} \rbf{± 0.0} & \textbf{70.3} \rbf{± 3.8} & \textbf{93.7} \rbf{± 3.2} & \underline{66.0} \rbf{± 7.5}  \\ 
        \bottomrule
    \end{tabular}}
    \caption{\textbf{Evaluation on RoboTwin benchmark.} We report the mean and standard deviation of success rates averaged over 3 random seeds.  Best score in \textbf{bold}, second-best \underline{underlined}. Our method outperforms the previous state-of-the-art with an average elevation of 11.4\% success rate across 16 tasks. - denotes that color-specific tasks are not suitable for DP3 while only taking point cloud as input.} 
    \label{tab: robotwin_eval}
    \vspace{-3mm}
\end{table*}

\noindent\textbf{Evaluation details.}
We evaluate each method with 3 random training seeds. For each seed, we conduct evaluations on 100 episodes (corresponding to 100 environment initializations) every 1,000 training steps and compute the average of the highest 3 success rates. We report the mean of success rates across 3 seeds. 


\noindent\textbf{Experimental results.} As shown in~\cref{tab:aloha_results}, our method achieves an average success rate of 71.9\%, surpassing all baselines in both tasks. This result demonstrates the superiority of the proposed approach in fine-grained bimanual manipulation. Furthermore, our method consistently outperforms baselines on both scripted and human-collected demonstrations, demonstrating its strong capability to capture the multi-modal nature of human demonstrations.

\subsubsection{Results on RoboTwin benchmark}

\noindent\textbf{RoboTwin benchmark.} RoboTwin benchmark \cite{mu2024robotwin} is built on ManiSkill~\cite{gu2023maniskill2}and features more complex scenarios, designed to evaluate the model's performance in terms of position generalization, visual distraction, and instance generalization. It consists of 16 diverse tasks, including \textit{put apple cabinet} and \textit{blocks stack}. Expert demonstrations are generated via 3D generative models and LLMs, enabling diverse and realistic task variations. Each task includes 100 demonstrations, spanning 250 to 850 steps (5 to 17 s). 

\noindent\textbf{Evaluation details.} We use the last checkpoint for evaluation on 100 episodes and report the mean and standard deviation of success rates across 3 random seeds \footnote{Given that both methods (GR-MG \cite{li2025gr_mg} and RDT-1B \cite{liu2024rdt}) are language-conditioned and applicable to multi-task settings, we fine-tune them separately on each task to ensure a fair evaluation.}. 


\begin{table}[t]
    \centering
    \resizebox{\linewidth}{!}{
    \begin{tabular}{lccc}
    \toprule
         Method & Dominant-select &  Sync-bimanual   & Seq-coordinate \\
         \midrule
         DP3~\cite{ze20243d_dp} & 56.7 & 40.6 & 29.0 \\
         DP~\cite{chi2023diffusion} & 32.0 & 40.0 & 15.4\\
         ACT~\cite{zhao2023aloha} & 51.2  & 43.6 & \underline{39.4} \\
          RDT-1B ~\cite{liu2024rdt} & \textbf{67.8}  & \underline{45.1} & 38.5 \\
         Ours & \underline{66.3}  & \textbf{49.4} & \textbf{50.8} \\
         \bottomrule
    \end{tabular}}
    \caption{\textbf{Performance comparison across bimanual task categories in the RoboTwin benchmark.} We categorize those tasks into 3 bimanual task types: (1) \textit{Dominant-select} – the executing arm is chosen based on the object’s position; (2) \textit{Sync-bimanual} – both arms operate independently but simultaneously; (3) \textit{Seq-coordinate} – tasks require sequential coordination with temporal dependencies between arms.}
    \label{tab: Result on different task categories.}
    \vspace{-3mm}
\end{table}

\noindent\textbf{Experimental results.} With quantitative results presented in~\cref{tab: robotwin_eval}, our method achieves an average success rate of 56.3\% across 16 bimanual manipulation tasks, making a significant absolute improvement of +14.0\% over 3D Diffusion Policy, +11.4\% over ACT and +5.2\% over RDT-1B. Notably, our method demonstrates substantial gains in long-horizon coordination tasks, such as \textit{block handover}, and \textit{put apple cabinet}, showcasing its ability to model complex sequential dependencies and enhance bimanual coordination. Additionally, our method exhibits strong robustness to visual distractions with superior performance in \textit{pick apple messy}. These results further validate the robustness of our method in cluttered environments.~\cref{tab: Result on different task categories.} provides a detailed performance comparison across 3 bimanual task categories in RoboTwin. Among them, all methods exhibit the lowest success rates on \textit{Seq-coordinate} tasks, highlighting the difficulty of precise long-horizon planning and temporal coordination. Our method achieves the highest success rate across all categories, outperforming others by a large margin, particularly in \textit{Seq-coordinate} tasks, where effective future prediction and unidirectional attention mechanisms enable superior sequential coordination.



\subsection{Ablation Study}
\label{subsec:Ablation Study}
We systematically conduct ablation studies to analyze the contribution of different components in our model and their impact on overall performance.

\begin{figure}[t]
    \centering
    \begin{subfigure}[b]{0.49\linewidth}
        \centering
        \resizebox{\linewidth}{!}{
        \input{tikz_figures/ablation1}}
        \caption{Different benchmarks.}\label{fig:ablation_video_prediction:a}
    \end{subfigure}
    \hfill
    \begin{subfigure}[b]{0.49\linewidth}
        \centering
        \resizebox{\linewidth}{!}{
        \input{tikz_figures/ablation2}}
        \caption{Different task categories.}\label{fig:ablation_video_prediction:b}
    \end{subfigure}
    \caption{\textbf{Effectiveness of video prediction}. Video prediction consistently improves performance across multiple benchmarks, highlighting its effectiveness. For data efficiency setting, each task in RoboTwin is provided with only 20 demonstrations. } 
    \label{fig:ablation_video_prediction}
    \vspace{-3mm}
\end{figure}

\noindent\textbf{Impact of future prediction.} To assess the contribution of video prediction, we compare our full model with a variant without video prediction. As shown in ~\cref{fig:ablation_video_prediction} (a), our method consistently outperforms the ablated variant across all benchmarks, confirming the effectiveness of video prediction as a key component in improving task success rates. To further analyze its impact across tasks with varying coordination requirements,  we categorize the 20 tasks into \textit{Sync-Bimanual}, \textit{Dominant-Select}, and \textit{Seq-Coordinated}, as shown in ~\cref{fig:ablation_video_prediction} (b). In \textit{Seq-Coordinated} tasks, e.g.,\textit{ handover}, video prediction achieves a substantial 6.2 \% improvement. By incorporating future prediction as an auxiliary task during training, the model learns better temporal dependencies, which improve the alignment of transitions between arm actions during execution. In contrast, \textit{Sync-Bimanua}l tasks, e.g., \textit{dual-object picking}, exhibit a marginal 0.7 \% decline, likely due to predictive modeling introducing unnecessary dependencies that interfere with independent execution. These results underscore the utility of video prediction in sequential coordination by reinforcing temporal reasoning and facilitating implicit coordination.

\begin{table}
  \centering
  \resizebox{\linewidth}{!}{
  \begin{tabular}{@{}ccl@{}}
    \toprule
    Future Prediction Type  & Video Representation &  Avg. SR. $ (\%) \uparrow$  \\
    \midrule
    -  &  - &   64.6 \\
    Next Frame & Pixels &  68.7 \gbf{+4.1}\\
    Next Frame &  Latents&  71.3 \gbf{+6.7} \\
    Multi-Frame &  Latents &  \textbf{71.9 \gbf{+7.3}}\\
    \bottomrule
  \end{tabular}}
  \caption{\textbf{Ablation of prediction strategy}. We compare variants with different prediction strategies with our baseline method (ours w/o prediction) on ALOHA benchmark. Our method benefits from multi-frame prediction and the use of future latent representations. }  
  \label{tab:prediction_type_results}
  \vspace{-3mm}
\end{table}

\noindent\textbf{Choice of future prediction strategy.} We evaluate the impact of different future prediction strategies, comparing representation types (raw pixels vs. latent features) and prediction targets (next-frame vs. multi-frame prediction). Results show that latent features provide a more compact and high-level representation, capturing task-relevant semantics while reducing redundancy, leading to improved performance over raw pixels. Additionally, multi-frame prediction outperforms next-frame prediction by enabling model to capture long-term task dependencies and anticipate future states, which is essential for effective action planning. Most importantly, regardless of the specific implementation, future prediction consistently improves performance. This may stem from its ability to provide better inductive bias, as future prediction encourages model to focus on dynamic information, which is beneficial across all tasks.

\begin{table}
  \centering
  \resizebox{\linewidth}{!}{
  \begin{tabular}{@{}lcc@{}}
    \toprule
    Attention Mechanism  &  Avg. SR.  $ (\%) \uparrow$  & Inference   (s) $\downarrow$\\
    \midrule
    Full attention  &  70.8  & 0.06 \\
    Decoupled attention &  70.1 & 0.03 \\
    Unidirectional attention (ours) & \textbf{71.9}  & \textbf{0.03} \\ \midrule
    GR-MG~\cite{li2025gr_mg} & 2.5 & 0.19 \\
    \bottomrule
  \end{tabular}}
  \caption{\textbf{Ablation of attention mechanism}.  Results show that our proposed unidirectional attention effectively balances task performance and inference efficiency. SR. means success rate.}
  \label{tab:attention_design_results}
   \vspace{-3mm}
\end{table}

\noindent\textbf{Choice of attention mechanism.} As illustrated in ~\cref{fig:attention_mechanism}, we compare different attention mechanisms to assess their impact on success rate and inference efficiency. Unidirectional attention achieves the highest success rate and the lowest inference time\footnote{Inference time is measured on an NVIDIA V100 GPU as the time taken to generate corresponding actions once an observation is received.}. By allowing future tokens to attend exclusively to past actions,  unidirectional attention effectively captures temporal dependencies between executed actions and visual outcomes. In contrast, full attention introduces unnecessary computational overhead due to redundant dependencies, while the decoupled variant underperforms because it restricts the influence of future predictions on action fitting, thereby reducing the model's ability to leverage environment dynamics for policy improvement. Furthermore, unidirectional attention eliminates the need for future prediction during inference, lowering computational costs and improving inference speed. Additionally, We further compare with GR-MG \cite{li2025gr_mg}, another video-prediction-based method, achieving significant gains in both inference speed and performance.

\noindent\textbf{Impact of video prediction weight.} We explored the effect of video prediction weight $\omega$ on ALOHA benchmark and the results are shown in Table~\ref{tab:albation_results_pred_weight}. The weights of video prediction are set to 0.05, 0.2, and 0.5. Among these,  $\omega = 0.2 $ achieved the best performance, suggesting that it effectively balances action fitting and video prediction. A lower weight ($\omega = 0.05$) might lead to insufficient guidance from video prediction, while a higher weight ($\omega = 0.5$) could overly constrain the model, hindering action optimization.

\begin{table}
    \centering
    \resizebox{\linewidth}{!}{
    \begin{tabular}{@{}l c c c c c@{}}
        \toprule
        Pred. & \cellcolor{gray!30} Avg. (\%) &  \multicolumn{2}{c}{Transfer Cube} & \multicolumn{2}{c}{Insertion}  \\
        Weight  & \cellcolor{gray!30} Success $\uparrow$ & Scripted & Human & Scripted & Human \\
        \midrule
        $\omega = 0.05$ & \cellcolor{gray!30} 71.1 & 95.5 & 78.3 & 83.5 & 26.9  \\
        $\omega = 0.2$  & \cellcolor{gray!30} \textbf{71.9} & 95.9 & 78.1 & 83.2 & 30.2  \\
        $\omega = 0.5$  & \cellcolor{gray!30} 70.6 & 95.4 & 78   & 82.2 & 26.9  \\
        \bottomrule
    \end{tabular}}
    \caption{\textbf{Ablation of video prediction weight.} $\omega = 0.2$ achieved the best result, effectively balancing action and video prediction.}
    \label{tab:albation_results_pred_weight}
     \vspace{-3mm}
\end{table}

\begin{table}
    \centering
    \resizebox{\linewidth}{!}{
    \begin{tabular}{@{}c c c l@{}}
        \toprule
        Chunk Size & \# Future Frames  & \# Visual Tokens & Avg. SR $\uparrow$ \\
        \midrule
        \multirow{4}{*}{100}
        & -   & -   &  64.6 \\
        & 20  & 192 & 71.1 \textcolor[rgb]{0.3, 0.7, 0.3}{ (+6.5)} \\
        & 40  & 396 & \textbf{71.9} \textcolor[rgb]{0.3, 0.7, 0.3}{ (+7.3)} \\
        & 100 & 960 & 70.6 \textcolor[rgb]{0.3, 0.7, 0.3}{ (+6.0)} \\
        \bottomrule
    \end{tabular}}
    \caption{\textbf{Impact of future frame number on ALOHA}. A moderate value led to optimal performance.}
    \label{tab:pred_len}
\end{table}

\noindent\textbf{Impact of video prediction horizon.} Visual tokens typically far outnumber action tokens. As previously discussed, we adopt strategies such as temporal downsampling, video tokenization, and patchification to control the visual token count. In this ablation, we further investigate how the length of the predicted future horizon affects performance. As shown in Table~\ref{tab:pred_len}, increasing the number of future frames from 20 to 40 yields better results, while extending it to 100 leads to a performance drop. The best outcome is achieved at 40 frames, indicating a balanced prediction horizon.

\noindent\textbf{Choice of chunk size.}  
\begin{figure}
    \centering
    \begin{subfigure}[t]{0.48\linewidth}
        \centering
        \resizebox{\linewidth}{!}{
        \input{tikz_figures/chunk1}}
        \caption{Effect of chunk size on ALOHA~\cite{zhao2023aloha} benchmark.}\label{fig:chunk_size:a}
    \end{subfigure}
    \hfill
    \begin{subfigure}[t]{0.49\linewidth}
        \centering
        \resizebox{\linewidth}{!}{
        \input{tikz_figures/chunk2}}
        \caption{Task-specific chunk size impact on RoboTwin~\cite{mu2024robotwin}  benchmark.}\label{fig:chunk_size:b}
    \end{subfigure}

    \caption{\textbf{The impact of chunk size on performance}.The optimal chunk size depends on the task, but video prediction consistently improves performance across all chunk sizes.}
    \label{fig:chunk_size}
     \vspace{-3mm}
\end{figure}
As shown in~\cref{fig:chunk_size} (a), a moderate chunk size yields the optimal performance, as high-frequency tasks benefit from action chunking, improving control stability. However, longer chunks limit the robot's capacity for adaptive response while increasing the complexity of modeling long action sequences, ultimately leading to performance degradation. This trend is consistent with previous studies \cite{zhao2023aloha, zhang2024arp}, highlighting the trade-off between action smoothness and reactivity. Notably, regardless of chunk size, video prediction consistently improves performance, highlighting its effectiveness in enhancing task success rates through the joint prediction of future frames and actions. ~\cref{fig:chunk_size} (b) indicates that optimal chunk size depends on task characteristics. Coordination tasks (e.g., \textit{blocks stack}) benefit from larger chunks, ensuring smoother collaboration when some tasks prefer smaller chunks. 



\subsection{Real-World Evaluation}

\begin{figure}
    \centering
    \includegraphics[width= 0.95\linewidth]{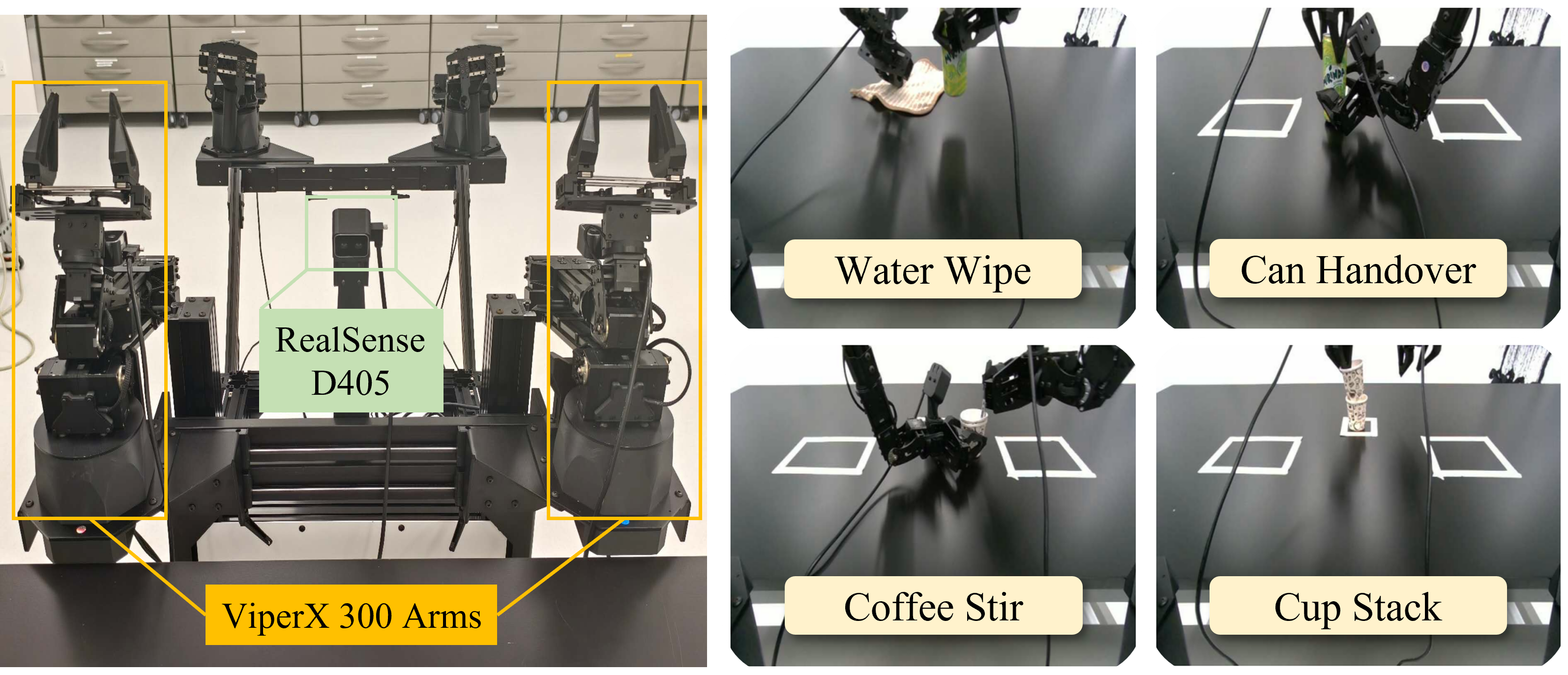}
    \caption{\textbf{Real-Robot setup and  4 real-world bimanual tasks.} We collect 50 demos for each task by human teleoperation.}
    \label{fig:real_robot}
     \vspace{-3mm}
\end{figure}

\noindent\textbf{Real-world benchmark.} We evaluate the methods on the Mobile Trossen robot, a variant of Mobile ALOHA~\cite{fu2024mobile_aloha}, which features two ViperX 300 as follower arms and two WidowX 250 as leader arms. The system captures RGB images at a resolution of 480×640 with a frame rate of 50 FPS. The benchmark consists of 4 tasks, including \textit{water wipe} and \textit{can handover}, covering a range of fine-grained bimanual manipulation scenarios. We collected 50 demonstrations for each task by human teleoperation. ~\cref{fig:real_robot} illustrates our real-robot setup and bimanual manipulation tasks.

\label{subsec:Real-world Evaluation}

\begin{table}[]
    \centering
    \resizebox{0.84\linewidth}{!}{
    \begin{tabular}{lccc}
    \toprule
      Task &  ACT \cite{zhao2023aloha}   &  DP \cite{chi2023diffusion} & Ours \\
       \midrule
       Water Wipe  &  30 & 10 & 40  \\
       Cup Stack &   40 & 0 & 70 \\
       Can handover & 20 & 10 & 70 \\
       Coffee Stir & 20 & 10 & 20 \\
       \rowcolor{gray!20}
     Avg. Success (\%) $\uparrow$  & 27.5 & 7.5 & \textbf{60.0}  \\
     \rowcolor{gray!20}
    Control Freq. (Hz)  $\uparrow$  & 33.9 &15.4 &\textbf{35.8}  \\
      \bottomrule   
    \end{tabular}}
  \caption{\textbf{Results of real-world experiments}. Our method outperforms all baselines in task success rate while maintaining high control frequency.}
  \label{tab:real_world}
   \vspace{-3mm}
\end{table}

\noindent\textbf{Evaluation details.}
Each method is evaluated over 10 episodes per task using the last checkpoint, and we report the success rate.

\noindent\textbf{Evaluation results and control frequency.}
\cref{tab:real_world} presents the results of real-world experiments. Our approach consistently achieves the best performance, demonstrating its effectiveness in real-world robotic experiments. All methods struggle with the \textit{Coffee Stir} task, which involves precise rotation, making it particularly challenging. We evaluate the control frequency for all methods with the same chunk size on an NVIDIA GeForce RTX 4060.Our method achieves the highest control frequency benefiting from a unidirectional attention mechanism that eliminates the dependency on future frame tokens during inference. This design ensures low-latency execution and efficient real-time control. 

\section{Conclusion}
This work presents a unified diffusion-based framework for joint video prediction and action fitting, optimizing the perception-prediction-control procedure in an end-to-end manner. Through a multi-frame latent prediction strategy and a unidirectional attention mechanism, our method effectively leverages predictive foresight to enhance bimanual coordination. Experimental results demonstrate consistent improvements over state-of-the-art methods across simulated and real-world benchmarks. While our approach significantly improves coordination and control efficiency, future work could explore incorporating multimodal sensory feedback and adaptive planning to enhance robustness in dynamic and unstructured environments. 

\noindent \textbf{Limitations and future work:} While our method demonstrates strong performance, several directions remain open. It may benefit from stronger visual encoders to handle fine-grained tasks involving small or occluded objects (e.g., pens). Slight drops in loosely coupled tasks suggest future exploration of arm-specific representations to avoid unnecessary dependencies. Lastly, selecting more informative keyframes as predictive targets instead of uniform sampling may enhance temporal reasoning and coordination.


%% file: tikz_figures/ablation1.tex
\begin{tikzpicture}

\definecolor{darkgray176}{RGB}{176,176,176}
\definecolor{darkseagreen11417699}{RGB}{114,176,99}
\definecolor{darkslateblue7495126}{RGB}{74,95,126}
\definecolor{lightgray204}{RGB}{204,204,204}

\begin{axis}[
legend columns=-1, 
legend style={fill=none,
      at={(0.5, 1.05)},
      anchor=south,
      font=\Large,
      draw opacity=1, text opacity=1, draw=black,
      /tikz/every even column/.append style={column sep=6pt},
      },
tick align=outside,
tick pos=left,
x grid style={darkgray176},
xmin=-0.485, xmax=2.485,
tick align=inside,
xtick={0,1,2},
xticklabels={ALOHA~\cite{zhao2023aloha},RoboTwin~\cite{mu2024robotwin},Data Efficiency},
xticklabel style={font=\Large, rotate=15},
ylabel={\Large Success Rate (\%)},
ymin=20, ymax=76.5,
ytick style={color=black}
]
\draw[draw=none,fill=darkseagreen11417699,line width=0.096pt,postaction={pattern=north east lines}] (axis cs:-0.35,0) rectangle (axis cs:0,64.6);
\addlegendimage{ybar,area legend,draw=none,fill=darkseagreen11417699,line width=0.096pt,postaction={pattern=north east lines}}
\addlegendentry{Ours w/o video prediction}

\draw[draw=none,fill=darkseagreen11417699,line width=0.096pt,postaction={pattern=north east lines}] (axis cs:0.65,0) rectangle (axis cs:1,53.6);
\draw[draw=none,fill=darkseagreen11417699,line width=0.096pt,postaction={pattern=north east lines}] (axis cs:1.65,0) rectangle (axis cs:2,25.2);
\draw[draw=none,fill=darkslateblue7495126,line width=0.096pt,postaction={pattern=north west lines}] (axis cs:2.77555756156289e-17,0) rectangle (axis cs:0.35,71.9);
\addlegendimage{ybar,area legend,draw=none,fill=darkslateblue7495126,line width=0.096pt,postaction={pattern=north west lines}}
\addlegendentry{Ours}

\draw[draw=none,fill=darkslateblue7495126,line width=0.096pt,postaction={pattern=north west lines}] (axis cs:1,0) rectangle (axis cs:1.35,56.3);
\draw[draw=none,fill=darkslateblue7495126,line width=0.096pt,postaction={pattern=north west lines}] (axis cs:2,0) rectangle (axis cs:2.35,27.3);
\draw (axis cs:0.175,72.9) node[
  scale=0.6,
  anchor=base,
  text=black,
  rotate=0.0
]{\LARGE\bfseries +7.3\%};
\draw (axis cs:1.175,57.3) node[
  scale=0.6,
  anchor=base,
  text=black,
  rotate=0.0
]{\LARGE\bfseries +2.7\%};
\draw (axis cs:2.175,28.3) node[
  scale=0.6,
  anchor=base,
  text=black,
  rotate=0.0
]{\LARGE\bfseries +2.1\%};
\end{axis}

\end{tikzpicture}

%% file: tikz_figures/ablation2.tex
\begin{tikzpicture}[]

\definecolor{darkgray176}{RGB}{176,176,176}
\definecolor{darkseagreen11417699}{RGB}{114,176,99}
\definecolor{darkslateblue7495126}{RGB}{74,95,126}
\definecolor{lightgray204}{RGB}{204,204,204}

\begin{axis}[
legend columns=-1, 
legend style={fill=none,
      at={(0.5, 1.05)},
      anchor=south,
      font=\Large,
      draw opacity=1, text opacity=1, draw=black,
      /tikz/every even column/.append style={column sep=6pt},
      },
tick align=outside,
tick pos=left,
x grid style={darkgray176},
xmin=-0.485, xmax=2.485,
xtick style={color=black},
xtick={0,1,2},
xticklabels={Sync-Bimanual,Dominant-Select,Seq-Coordinated},
y grid style={darkgray176},
xticklabel style={font=\Large, rotate=15},
ylabel={\Large Success Rate (\%)},
ymin=45, ymax=69,
tick align=inside,
ytick style={color=black}
]
\draw[draw=none,fill=darkseagreen11417699,line width=0.096pt,postaction={pattern=north east lines}] (axis cs:-0.35,0) rectangle (axis cs:0,50.08);
\addlegendimage{ybar,area legend,draw=none,fill=darkseagreen11417699,line width=0.096pt,postaction={pattern=north east lines}}
\addlegendentry{Ours w/o video prediction}

\draw[draw=none,fill=darkseagreen11417699,line width=0.096pt,postaction={pattern=north east lines}] (axis cs:0.65,0) rectangle (axis cs:1,64);
\draw[draw=none,fill=darkseagreen11417699,line width=0.096pt,postaction={pattern=north east lines}] (axis cs:1.65,0) rectangle (axis cs:2,53.26);
\draw[draw=none,fill=darkslateblue7495126,line width=0.096pt,postaction={pattern=north west lines}] (axis cs:2.77555756156289e-17,0) rectangle (axis cs:0.35,49.42);
\addlegendimage{ybar,area legend,draw=none,fill=darkslateblue7495126,line width=0.096pt,postaction={pattern=north west lines}}
\addlegendentry{Ours}

\draw[draw=none,fill=darkslateblue7495126,line width=0.096pt,postaction={pattern=north west lines}] (axis cs:1,0) rectangle (axis cs:1.35,66.33);
\draw[draw=none,fill=darkslateblue7495126,line width=0.096pt,postaction={pattern=north west lines}] (axis cs:2,0) rectangle (axis cs:2.35,59.42);
\draw (axis cs:0.175,49.92) node[
  scale=0.6,
  anchor=base,
  text=black,
  rotate=0.0
]{\LARGE\bfseries -0.7\%};
\draw (axis cs:1.175,66.83) node[
  scale=0.6,
  anchor=base,
  text=black,
  rotate=0.0
]{\LARGE\bfseries +2.3\%};
\draw (axis cs:2.175,59.92) node[
  scale=0.6,
  anchor=base,
  text=black,
  rotate=0.0
]{\LARGE\bfseries +6.2\%};
\end{axis}

\end{tikzpicture}

%% file: tikz_figures/chunk1.tex
\begin{tikzpicture}

\definecolor{darkgray176}{RGB}{176,176,176}
\definecolor{darkseagreen11417699}{RGB}{114,176,99}
\definecolor{darkslateblue7495126}{RGB}{74,95,126}
\definecolor{lightgray204}{RGB}{204,204,204}

\begin{axis}[
legend columns=-1, 
legend style={fill=none,
      at={(0.5, 1.05)},
      anchor=south,
      font=\large,
      draw opacity=1, text opacity=1, draw=black,
      /tikz/every even column/.append style={column sep=6pt},
      },
tick align=inside,
tick pos=left,
x grid style={darkgray176},
xmin=0.7, xmax=7.3,
xtick style={color=black},
xtick={1,2,3,4,5,6,7},
xticklabels={20,40,60,80,100,200,400},
y grid style={darkgray176},
ylabel={\Large Success Rate (\%)},
ymajorgrids,
ymin=50, ymax=73,
ytick style={color=black}
]
\addplot [thick, darkseagreen11417699, dashed, mark=square*, mark size=3, mark options={solid}]
table [row sep=\\] {
1 54.96 \\
2 63.25 \\
3 63.5 \\
4 63.92 \\
5 64.64 \\
6 64.25 \\
7 61.08 \\
};
\addlegendentry{Ours w/o video prediction}
\addplot [thick, darkslateblue7495126, mark=*, mark size=3, mark options={solid}]
table [row sep=\\] {
1 56.25 \\
2 64.92 \\
3 67 \\
4 67.92 \\
5 71.86 \\
6 66.75 \\
7 65.5 \\
};
\addlegendentry{Ours}
\end{axis}

\end{tikzpicture}

%% file: tikz_figures/chunk2.tex
\begin{tikzpicture}

\definecolor{darkgray176}{RGB}{176,176,176}
\definecolor{darkseagreen11417699}{RGB}{114,176,99}
\definecolor{darkslateblue7495126}{RGB}{74,95,126}
\definecolor{lightgray204}{RGB}{204,204,204}
\definecolor{silver184219179}{RGB}{184,219,179}

\begin{axis}[
legend columns=-1, 
legend style={fill=none,
      at={(0.5, 1.05)},
      anchor=south,
      font=\large,
      draw opacity=1, text opacity=1, draw=black,
      /tikz/every even column/.append style={column sep=6pt},
      },
tick align=inside,
tick pos=left,
x grid style={darkgray176},
xmin=-0.48, xmax=3.48,
xtick style={color=black},
xtick={0,1,2,3},
xticklabels={
  Dual Bottles\\Pick (Hard),
  Blocks Stack Easy,
  Bottles Adjust,
  Empty Cup Place
},
xticklabel style={text width=5em, align=center},
y grid style={darkgray176},
ylabel={\Large Success Rate (\%)},
ymin=0, ymax=77,
ytick style={color=black}
]
\draw[draw=none,fill=darkslateblue7495126,postaction={pattern=north east lines}] (axis cs:-0.3,0) rectangle (axis cs:-0.1,49);
\addlegendimage{ybar,area legend,draw=none,fill=darkslateblue7495126,postaction={pattern=north east lines}}
\addlegendentry{Chunk = 10}

\draw[draw=none,fill=darkslateblue7495126,postaction={pattern=north east lines}] (axis cs:0.7,0) rectangle (axis cs:0.9,28);
\draw[draw=none,fill=darkslateblue7495126,postaction={pattern=north east lines}] (axis cs:1.7,0) rectangle (axis cs:1.9,54);
\draw[draw=none,fill=darkslateblue7495126,postaction={pattern=north east lines}] (axis cs:2.7,0) rectangle (axis cs:2.9,69);
\draw[draw=none,fill=darkseagreen11417699,postaction={pattern=north west lines}] (axis cs:-0.1,0) rectangle (axis cs:0.1,53);
\addlegendimage{ybar,area legend,draw=none,fill=darkseagreen11417699,postaction={pattern=north west lines}}
\addlegendentry{Chunk = 20}

\draw[draw=none,fill=darkseagreen11417699,postaction={pattern=north west lines}] (axis cs:0.9,0) rectangle (axis cs:1.1,60);
\draw[draw=none,fill=darkseagreen11417699,postaction={pattern=north west lines}] (axis cs:1.9,0) rectangle (axis cs:2.1,57);
\draw[draw=none,fill=darkseagreen11417699,postaction={pattern=north west lines}] (axis cs:2.9,0) rectangle (axis cs:3.1,65);
\draw[draw=none,fill=silver184219179,postaction={pattern=crosshatch}] (axis cs:0.1,0) rectangle (axis cs:0.3,51);
\addlegendimage{ybar,area legend,draw=none,fill=silver184219179,postaction={pattern=crosshatch}}
\addlegendentry{Chunk = 40}

\draw[draw=none,fill=silver184219179,postaction={pattern=crosshatch}] (axis cs:1.1,0) rectangle (axis cs:1.3,71);
\draw[draw=none,fill=silver184219179,postaction={pattern=crosshatch}] (axis cs:2.1,0) rectangle (axis cs:2.3,59);
\draw[draw=none,fill=silver184219179,postaction={pattern=crosshatch}] (axis cs:3.1,0) rectangle (axis cs:3.3,8);
\end{axis}

\end{tikzpicture}